\documentclass{bjmc}   

\usepackage{natbib}       

\usepackage{amsmath}
\usepackage{amssymb} 

\usepackage{array}
\usepackage[dvipsnames]{xcolor}
\usepackage{float}
\usepackage{float}

\usepackage{tikz}

\usepackage[hidelinks]{hyperref}

\usepackage{graphicx}
\usepackage{subcaption}
\usepackage{caption}

\usepackage{booktabs}
\usepackage{threeparttable}

\usepackage[dvipsnames]{xcolor}

\usepackage{algorithm2e}

\usepackage{listings}
\usepackage{xcolor}

\definecolor{codegreen}{rgb}{0,0.6,0}
\definecolor{codegray}{rgb}{0.5,0.5,0.5}
\definecolor{codepurple}{rgb}{0.58,0,0.82}
\definecolor{backcolour}{rgb}{0.95,0.95,0.92}

\lstdefinestyle{mystyle}{
    backgroundcolor=\color{backcolour},   
    commentstyle=\color{codegreen},
    keywordstyle=\color{magenta},
    numberstyle=\tiny\color{codegray},
    stringstyle=\color{codepurple},
    basicstyle=\ttfamily\footnotesize,
    breakatwhitespace=false,         
    breaklines=true,                 
    captionpos=b,                    
    keepspaces=true,                 
    numbers=left,                    
    numbersep=5pt,                  
    showspaces=false,                
    showstringspaces=false,
    showtabs=false,                  
    tabsize=2
}

\usepackage{multirow}
\usepackage{verbatim}

\begin{document}

%


\title{%
  \texorpdfstring
    {GFLC: Graph‐based Fairness‐aware\\Label Correction for Fair Classification}
    {GFLC: Graph-based Fairness-aware Label Correction for Fair Classification}
}



\titlerunning{GFLC: Graph-based Fairness-aware Label Correction}  
%
%
\author{Modar SULAIMAN, Kallol ROY
}
\authorrunning{Sulaiman et al.}   
%
%
\institute{University of Tartu, Institute of Computer Science, Tartu, Estonia
} 
%
\emails{modar.sulaiman@ut.ee, kallol.roy@ut.ee}
\orcids{ORCID: 0000-0002-7322-078X, ORCID: 0000-0002-6557-2689}
%
%

\maketitle      

\begin{abstract}

Fairness in machine learning (ML) has a critical importance for building trustworthy machine learning system as artificial intelligence (AI) systems increasingly impact various aspects of society, including healthcare decisions and legal judgments. Moreover, numerous studies demonstrate evidence of unfair outcomes in ML and the need for more robust fairness-aware methods. However, the data we use to train and develop debiasing techniques often contains biased and noisy labels. As a result, the label bias in the training data affects model performance and misrepresents the fairness of classifiers during testing. To tackle this problem, our paper presents Graph-based Fairness-aware Label Correction (GFLC), an efficient method for correcting label noise while preserving demographic parity in datasets. In particular, our approach combines three key components: prediction confidence measure, graph-based regularization through Ricci-flow-optimized graph Laplacians, and explicit demographic parity incentives. Our experimental findings show the effectiveness of our proposed approach and show significant improvements in the trade-off between performance and fairness metrics compared to the baseline.\footnote{Preprint: The code used for conducting the experiments will be published publicly once the paper is accepted.}

\end{abstract}

\section{Introduction}

Machine learning has become widely used in different critical real-world applications that significantly impact society, leading to growing concerns about fairness in automatic decision-making systems. Research has shown that, without appropriate intervention, these models can exhibit bias against specific demographic groups \citep{dwork2012fairness, hardt2016equality}. Therefore, research on fair classification has garnered significant attention, leading to the development of various methods aimed at improving fairness \citep{kearns2018preventing, zafar2017fairness, mehrabi2021survey, sulaiman2025advancing}. These methods typically focus on specific notions of fairness and often assume that predefined sensitive information is available during the training process. However, most methods rely on the assumption that labels are completely accurate. This is challenging to achieve because of how these labels are generated \citep{wang2021fair}. For example, the ImageNet-scale dataset was annotated by various distributed workers on Amazon Mechanical Turk\footnote{\url{https://www.mturk.com}} \citep{han2020survey, wu2022fair}. Moreover, according to \citep{northcutt2021pervasive, wu2022fair}, the test sets of ten widely used computer vision, natural language, and audio datasets contain an average label noise of 3.4\%.

In the research domain of learning with noisy labels, label noise can be categorized into different types \citep{algan2021image}. For example, assume that \( x \) represents the instance features, \( s \) denotes the sensitive group, and \( y \) is the class label, we have the following cases: (i) Random noise: This type is completely random and does not depend on the instance features (\( x \) or/and \( s \)) or the true class label \( y \). (ii) Y-dependent noise: This type is independent of the instance features (\( x \) and \( s \)) but is influenced by the class label \( y \). (iii) XY-dependent noise: This type depends on both the instance features (\( x \) and \( s \)) and the class label \( y \). Furthermore, when noise is specifically affected by a sensitive attribute $s$ and its corresponding class $y$ \citep{e2024fair}, it is referred to in the fairness literature as group-dependent \citep{wang2021fair} or instance-dependent \citep{wu2022fair}. This situation is related to discrimination and is the main focus of our paper. In such cases, the likelihood of an instance being noisy is influenced by both its class \( y \) and the sensitive group \( s \) to which it belongs. This suggests that label noise is particularly important when considering bias mitigation techniques.

Researchers \citep{lamy2019noise, liu2021can, wang2021fair, fogliato2020fairness} conducted experiments that show that enforcing parity constraints on noisy labels can negatively impact the classifier's accuracy for groups unaffected by label noise. This shows that label noise adversely affects the performance of machine learning models. Moreover, label noise significantly affects fairness metrics and can cause some fairness-aware algorithms to be more biased than those not designed with fairness in mind \citep{wang2021fair}. In our paper, we tackle the previous challenges that are related to learning with noisy labels and fairness in machine learning. Consequently, we present a novel approach, called Graph-based Fairness-aware Label Correction (GFLC). Our GFLC method is specifically designed to correct labels while ensuring fairness between different demographic groups. GFLC combines graph-based techniques, fairness-aware learning and prediction confidence measure, effectively utilizing the structural insights offered by k-nearest neighbor (k-NN) graphs, Forman–Ricci curvature, and discrete Ricci flow. In summary, our contributions are as follows.

\begin{itemize}
  
\item \textbf{GFLC:} We introduce Graph-based Fairness-aware Label Correction (GFLC) a novel method to address the problem of instance-dependent label noise in a biased dataset. GFLC is designed to enhance the trade-off between performance and demographic parity in the presence of noisy labels in biased datasets.

\item \textbf{Empirical Observation:} Experiments validate our claims for the proposed approach, demonstrating improvements in the trade-off between fairness and performance, especially at high noise rate.

\end{itemize}

The structure of this article is as follows: Section \ref{Related_work} presents related work relevant to our paper, while Section \ref{Preliminaries_sec} covers the preliminaries. Our proposed method, called GFLC, is described in Section \ref{section:GFLC}. Section \ref{Experimental_setup} outlines the experimental setup for our research. Section \ref{Results_sec} presents the evaluation of our GFLC method in comparison to the baseline method. Finally, Section \ref{sec_Conclusion} concludes our work.

\section{Related work}\label{Related_work}

In this section, we present relevant literature on fairness in machine learning when dealing with group-dependent noisy labels, Ricci curvature in network science, learning with noisy labels, graph Laplacian, and Semi-Supervised Learning (SSL).

\paragraph{Fairness in Machine Learning and Label Correction.} 

In our work, we primarily focus on addressing fairness issues in machine learning in the presence of biased and group-dependent noisy training labels. \citep{wang2021fair} presented a method for training fair classifiers in scenarios where training labels are subject to random noise where the error rates of corruption are influenced by both the label class and the protected attribute. Furthermore, \citep{wang2021fair} presented analytical results demonstrating that simply imposing parity constraints on demographic disparity measures, without considering varying error rates between different groups, can reduce both the performance and fairness of the resulting classifier. They address such issues by employing empirical risk minimization with addtion to a surrogate loss functions and constraints. This approach helps to mitigate the challenges posed by heterogeneous label noise.

The work by \citep{wu2022fair} introduced general frameworks for developing fair classifiers that consider instance-dependent label noise. In their study, they addressed statistical fairness by reformulating both the classification risk and fairness metrics in the context of noisy data, which allowed for the creation of more robust classifiers. For their causality-based approach to fairness, they utilized the internal causal structure of the data to simultaneously model label noise and ensure counterfactual fairness. In addition, \citep{tjandra2023leveraging} proposed an approach to address instance-dependent label noise without making assumptions about the noise distribution, utilizing all available data during training. Their focus is on a scenario commonly found in healthcare, where researchers are provided with observed labels for a condition of interest, such as cardiovascular disease. They assume that a clinical expert can evaluate the accuracy of these observed labels for a small subset of the data, for instance, through a manual chart review. Using this specific subset 'alignment set' \citep{tjandra2023leveraging} identifies the underlying patterns of label noise. They then minimize a weighted cross-entropy across all the data. The authors note that their alignment set is a particular instance of anchor points \citep{Liu2016}, with the additional requirement that it includes instances where the ground truth and observed labels either match or do not match.

\citep{canalli2024fair} explored the intersection of fair machine learning techniques and fundamental sources of bias, such as noisy data, emphasizing similarities and differences between fairness and noise in the context of machine learning. In addition, they developed the Fair Transition Loss, a new method for fair classification inspired by robust label noise learning techniques. Furthermore, the study presented by \citep{e2024fair} introduces Fair-OBNC, a novel method for correcting label noise that considers fairness. This approach builds upon the Ordering-Based Label Noise Correction (OBNC) algorithm \citep{feng2015class} by incorporating fairness factors. Fair-OBNC specifically aims to enhance demographic parity in training datasets, making it a distinctive method at the intersection of fairness and noise correction. Additionally, Fair-OBNC \citep{e2024fair} is the most closely related work to our paper, as it serves as the baseline and uses the same dataset for evaluation.

\paragraph{Ricci Curvature in Network Science.} In classical mathematics, Ricci curvature and Ricci flow are among the most important tools for analyzing manifolds according to their geometric and topological properties. \citep{sreejith2016forman} introduced a new discretization of the classical Ricci curvature, which was originally proposed by \citep{forman2003bochner}, specifically for the domain of complex networks. Furthermore, \citep{sreejith2016forman} explored the correlation between Forman curvature and common network measures, such as degree, clustering coefficient, and betweenness centrality, in both model and real networks. In work by \citep{weber2016forman}, they built upon and expanded the work by \citep{sreejith2016forman}. Their research focused on examining the relationship between Forman-Ricci curvature and other geometric properties of networks, such as node degree distribution and connectivity structure. Additionally, \citep{weber2016forman} introduced an innovative change detection method for complex dynamic networks, which leverages Ricci flow on the edges in relation to Forman curvature. In addition, in the work by \citep{weber2017characterizing}, they applied Forman's Ricci curvature and the Bochner Laplacian \citep{forman2003bochner} to analyze networks and explore their effectiveness as network characteristics. Additionally, they introduced a method for detecting changes in evolving networks and proposed the Laplacian flow as a tool for denoising networks constructed from empirical data. Furthermore, \citep{weber2017characterizing} noted in their study that denoising could be a practical application resulting from their theoretical findings. Besides, in an empirical investigation, \citep{samal2018comparative} examined two discretizations of Ricci curvature: Ollivier's Ricci curvature and Forman's Ricci curvature. They analyzed these measures across various model and real-world networks and found a strong correlation between those two types of Ricci curvature in many cases. Moreover, a significant advantage of Forman-Ricci curvature, as highlighted by \citep{samal2018comparative}, is that it captures essential geometric properties of networks while being much simpler to compute on large networks compared to Ollivier-Ricci curvature. Additionally, they noted that Forman-Ricci curvature can be used as a faster alternative to Ollivier-Ricci curvature for coarse analyses in larger real-world networks, when less precision is acceptable. 

An innovative geometric approach was introduced in \citep{ni2019community} that enables the use of powerful classical geometric methods and properties. By conceptualizing networks as geometric objects and considering communities within a network as geometric decompositions, they apply concepts such as curvature and discrete Ricci flow methods. These techniques have demonstrated significant effectiveness in analyzing and breaking down communities in networks. Moreover, \citep{topping2021understanding} examined the concept of graph bottlenecks and the issue of over-squashing, which limits the performance of message passing in graph neural networks. They introduced the Balanced Forman curvature, a new type of edge-based Ricci curvature that connects to the classical Ollivier curvature. Furthermore, the authors demonstrated that negatively curved edges contribute to the problem of over-squashing, suggesting that curvature-based rewiring of the graph could improve its bottleneck characteristics. In addition, the work by \citep{nguyen2023revisiting} established a new connection between Ollivier-Ricci curvature on graphs and the challenges of over-smoothing and over-squashing. They demonstrated that positive graph curvature is associated with over-smoothing, while negative graph curvature is linked to over-squashing. To address these issues, \citep{nguyen2023revisiting} proposed an innovative curvature-based rewiring method called Batch Ollivier-Ricci Flow (BORF), which effectively improves the performance of graph neural networks (GNNs) by tackling both over-smoothing and over-squashing simultaneously.

\paragraph{Graph Laplacian} Learning on graphs with Laplacian regularization has been explored by \citep{ando2006learning}. They derived generalization bounds for this approach, utilizing the properties of the graph. Furthermore, \citep{ando2006learning} emphasized the significance of Laplacian normalization and dimensionality reduction in graph learning. Moreover, \citep{streicher2023graph} proposed a new definition of the graph-Laplacian aimed at enhancing performance for Semi-Supervised Learning (SSL) tasks. Furthermore, the proposed operator in \citep{streicher2023graph} facilitates smooth interpolation between the unsupervised and semi-supervised scenarios. In addition, in \citep{de2020analysis}, the authors conducted a comprehensive empirical evaluation of different graph-based semi-supervised learning (SSL) algorithms for label noise. They demonstrated that detecting the noisiest instances is possible if the dataset is consistent with the assumptions of semi-supervised learning (SSL). Furthermore, their findings revealed that in high-dimensional clusters, the Laplacian eigenmaps (LE) algorithm outperformed label propagation. Additionally, the work by \citep{chen2025laplaceconfidence} introduced LaplaceConfidence, a method that obtains label confidence (clean probabilities) using Laplacian energy. They demonstrate that their approach outperforms other classification-loss-based estimation methods, achieving state-of-the-art results on standard benchmarks for Learning with Noisy Labels (LNL). In comparison to previous works, our current work addresses the issue of fairness in machine learning and examines how noisy labels impact fairness in classification tasks.

\section{Preliminaries}\label{Preliminaries_sec}

Let $G=(V,E,W)$ be an undirected graph with vertex set $V$, edge set $E$, and a positive weight $w_{uv}\in W$ assigned to each edge $e_{uv}=\{u,v\}\in E$.  We write $x\sim u$ to mean that vertex $x$ is a neighbor of $u$ (i.e.\ $\{u,x\}\in E$). 

\subsection{k-NN Graph Construction}\label{k-NN_graph}

A $k$-Nearest Neighbors (k\,NN) graph is a graph where each node (data point) is connected to its $k$ nearest neighbors based on a distance metric. For dataset \(\mathcal{D} = \{\mathbf{x}_i, \tilde{y}_i, s_i\}_{i=1}^N\),  we construct an undirected $k$-nearest-neighbor graph $G=(V,E)$  from features to capture the data manifold. Moreover, we define the inverse distance weighting for edge weights as follows:

\begin{equation}
w_{ij} = \frac{1}{\max(d(\mathbf{x}_i, \mathbf{x}_j), \epsilon)}   
\end{equation}

where $\epsilon = 10^{-8}$ is a small constant to prevent division by zero, and $d(\mathbf{x}_i, \mathbf{x}_j)$ is the Euclidean distance ($L_2$ norm) between nodes $i$ and $j$. Moreover,  the weights will be large when two points are similar (i.e. when the distance is small), and conversely small when they are dissimilar (i.e. when the distance is large).

\subsection{Forman–Ricci curvature}

\citep{sreejith2016forman} introduced the Forman curvature, which represents a new type of discretization of Ricci curvature, to the realm of undirected networks. In the combinatorial Forman–Ricci curvature framework, each edge $e_{uv}$ is assigned a curvature value $F(e_{uv})$ that reflects the local connectivity around $u$ and $v$. In general, it measures edge strength relative to neighborhood dispersion. The mathematical formula of the Forman curvature \citep{sreejith2016forman} (Equation \ref{eq:Forman–Ricci1}) elegantly incorporates the weights of nodes and edges in networks. Hence, Forman curvature can be utilized to study the structure of both unweighted and weighted complex networks.

\begin{equation}\label{eq:Forman–Ricci1}
    F(e_{uv}) = w_{uv} \left( \frac{w_u}{w_{uv}} + \frac{w_v}{w_{uv}} - \sum_{x \sim u} \frac{w_u}{\sqrt{w_{uv} w_{ux}}} - \sum_{x \sim v} \frac{w_v}{\sqrt{w_{uv} w_{vx}}} \right)
\end{equation}

Here the sums run over all neighbors $x$ of $u$ and $v$, respectively (typically excluding $v$ from the sum over $u$ and vice versa).  Intuitively, each term $\sqrt{w_{uv}/w_{ux}}$ compares the weight of edge $uv$ to the weight of an adjacent edge $ux$. This expression arises from Forman’s general discretization of Ricci curvature on CW-complexes \citep{forman2003bochner}, specialized to graphs \citep{sreejith2016forman, weber2017characterizing}. Moreover, Forman-Ricci curvature is better suited to investigate networks where edge weights correspond to distances \citep{samal2018comparative}.

\subsection{Simplified Forman–Ricci curvature}\label{Simplified_Forman}

Based on \citep{samal2018comparative}, where they considered combinatorial case, i.e. for $w_{e} = w_{v} = 1, e \in E(G), v \in V(G) $ where $E(G)$ and $V(G)$ represent the set of
edges and vertices, respectively, in graph $G$, we consider a simplified version of the previous Forman–Ricci curvature (Equation \ref{eq:Simplified-Forman–Ricci}). Notably, (Equation \ref{eq:Simplified-Forman–Ricci}) omits vertex weights \( w_u \) and \( w_v \), relying solely on edge weights. The derivation of this simplified form from the general formula can be found in the Appendix \ref{appendix_A}. In general, the simplified Forman curvature formula prioritizes edge-centric properties, measuring whether an edge acts as a bottleneck (negative curvature) or facilitates expansion (positive curvature). More precisely, for $e_{uv}=\{u,v\}\in E$, we define the discrete Forman-Ricci curvature $F(e_{uv})^{(t)}$ at iteration $t$ as follows:

\begin{equation}\label{eq:Simplified-Forman–Ricci}
F(e_{uv})^{(t)} = w_{uv}^{(t)} \left(1 - \frac{1}{2}\left(
\sum_{x \in \mathcal{N}(u) \setminus \{v\}}\sqrt{\frac{w_{uv}^{(t)}}{w_{ux}^{(t)}}} + 
\sum_{x \in \mathcal{N}(v) \setminus \{u\}}\sqrt{\frac{w_{uv}^{(t)}}{w_{vx}^{(t)}}}
\right)\right)
\end{equation}

where $w_{uv}^{(t)}$ is the weight of edge $(u,v)$ at iteration $t$,  $\frac{w_{uv}^{(t)}}{W_{ux}^{(t)}}$ is the ratio of current edge weight to neighbor edge weight from node $u$, and $\frac{w_{uv}^{(t)}}{w_{vx}^{(t)}}$ is ratio of current edge weight to neighbor edge weight from node $v$. Moreover, we denote the following:

\begin{equation}
\text{sum}_u = \underbrace{\sum_{x \sim u}\sqrt{\frac{w_{uv}^{(t)}}{w_{ux}^{(t)}}}}_{\text{Node u neighbors}}  \quad  \text{and}  \quad  \text{sum}_v = \underbrace{\sum_{x \sim v}\sqrt{\frac{w_{uv}^{(t)}}{w_{vx}^{(t)}}}}_{\text{Node v neighbors}}
\end{equation}

$$F(e_{uv})^{(t)}= w_{uv}^{(t)} \cdot \left(1 - \frac{\text{sum}_u + \text{sum}_v}{2}\right) = w_{uv}^{(t)} \cdot (1-A)$$

Next, we have the following cases based on the curvature sign \citep{ni2019community, topping2021understanding}: (i) \textit{Positive Curvature:} ($F(e_{uv})^{(t)} > 0$) occurs when $A < 1$ and indicates that the edge $(u,v)$ is relatively strong compared to its neighborhood. Moreover, it suggests good local connectivity and structural importance. (ii) \textit{Negative Curvature:} ($F(e_{uv})^{(t)} < 0$) occurs when $A > 1$ and indicates that the edge $(u,v)$ is relatively weak compared to its neighborhood. Moreover, it suggests a structural bottleneck or sparse connectivity (e.g., a critical connection in a network). (iii) \textit{Zero Curvature:} ($F(e_{uv})^{(t)} = 0$) occurs when $A= 1$. Moreover, this indicates balanced local connectivity.

\subsection{Ricci Flow Update Rule}

In Hamilton's original formulation \citep{hamilton1982three} the Ricci flow equation is written as:

\begin{equation}\label{Riici_eq_1}
\frac{\partial g_{ij}}{\partial t} = -2\text{Ric}_{ij}
\end{equation}

In addition, the negative sign in the previous Equation (\ref{Riici_eq_1}) indicates that for regions of positive Ricci curvature will have metric contraction, while regions of negative curvature undergo expansion. This convention aims to eliminate curvature concentrations and achieve geometric uniformity. Generally, Ricci flow is a geometric evolution equation that smooths the geometry of manifolds by evolving the metric tensor according to its Ricci curvature.

In the context of graphs, discrete Ricci flow will update each edge to enhance the graph's geometric properties and connectivity patterns. Furthermore, the discrete Ricci flow algorithm on a network is an iterative process in which we update all edge weights simultaneously in each iteration \citep{ni2019community}. In \citep{ni2019community}, a discrete version of this flow process was introduced to update all edge weights. However, translating from continuous manifolds to discrete graphs requires careful consideration of both the geometric interpretation and the desired optimization objectives. The choice of sign in Equation (\ref{Riici_eq_1}) fundamentally influences whether the algorithm enhances or diminishes connections based on curvature properties. In our paper, we consider adjusting and updating the edge weights for a limited number of iterations as follows:

\begin{equation}\label{weight_update_rule4}
w_{ij}^{(t+1)} = \max\left(w_{ij}^{(t)} + \eta \cdot F(e_{uv})^{(t)}, \quad \epsilon\right)
\end{equation}

In this context, $\eta = 0.1$ represents the flow step (learning rate). We use $\epsilon$, a small positive constant set to $1e-8$, to prevent weights from becoming zero. The expression $\max(\cdot, \epsilon)$ ensures that the weights remain positive. Moreover, based on Equation (\ref{weight_update_rule4}), we can derive the following geometric interpretation and discuss its evolutionary consequences: \textit{First:} Edges with \( F(e_{uv}) > 0 \) experience increases in weight. This amplification process, described in Equation (\ref{weight_update_rule4}), generates several cascading effects. For instance, as the weights increase, the effective connectivity between well-connected regions becomes stronger, resulting in more robust intra-community bonds. \textit{Second:} When $F(e_{uv}) < 0$, Equation (\ref{weight_update_rule4}) produces weight reductions. Furthermore, we can see that inter-community edges experience weight reductions which can effectively increasing the separation between distinct network modules. This process enhances community detection performance by amplifying modularity differences. Moreover, for the sparse tree-like regions we can obserave that they will experience weight reductions that effectively prune weak connections, encouraging the formation of more compact, well-connected components. This process mirrors the geometric flow's tendency to eliminate irregularities.

In summary, in Equation (\ref{weight_update_rule4}) we can observe that positively curved edges shrink, thereby increasing their weight, while negatively curved edges expand. Furthermore, the characteristics of Equation (\ref{weight_update_rule4}) contribute significantly to the main goal of our proposed GFLC method. Specifically, we aim to enhance the graph Laplacian term (see Section \ref{Graph_Laplacian}), which focuses on ranking or scoring data points that may contain noise.

\section{Our Prposed Method (GFLC)}\label{section:GFLC}

In this section, we introduce our method \textit{Graph-based Fairness-aware Label Correction} (GFLC). GFLC is a method that integrates prediction confidence, geometric structure via graph Laplacian and Ricci flow, and fairness via demographic parity for fairness-aware label noise correction. GFLC takes as input a labeled dataset with features $X=\{x_i\}_{i=1}^n$, (possibly noisy) labels $y=\{y_i\}\in\{0,1\}$ and a sensitive attribute $s=\{s_i\}\in\{0,1\}$ indicating group membership.  It outputs corrected labels $\tilde y$ that aim to improve fairness across groups while respecting the geometric structure of data and classifier confidence.

\paragraph{GFLC proceeds as follows:} \textit{First:} train a classifier on $(X,y)$ to obtain class probabilities $p_i=\Pr(y_i=1\mid x_i)$ (for computing the margin term); \textit{Second:} build a KNN graph (Section \ref{k-NN_graph}); \textit{Third:} compute discrete Forman–Ricci curvature on the edges and apply a Ricci flow rewiring to adjust the graph structure. \textit{Fourth:} compute the fairness term. \textit{Finally:} computes a \textit{Combined Correction Score} (\( score_i \)) that trades off margin term (prediction confidence), graph Laplacian regularization term, and a fairness penalty to determine which labels should be corrected at the end. The combined correction score for each data point $x_i$ is computed as:

\begin{equation}\label{scorei}
  \text{score}_i
    = \underbrace{\alpha(1 - M_i)}_{\text{Margin term}}
    + \underbrace{ \beta L_i}_{\text{Graph Laplacian}}
    + \underbrace{\gamma\,\Delta\mathrm{DP}_i}_{\text{Fairness incentive}}
\end{equation}

The resulting multi-component scoring function ensures high-quality corrections for each data point by scoring the instances using \( score_i \) in descending order, which reflects the most uncertain instances via the margin term, the weight via Graph Laplacian for instance \( x_i \), and the highest fairness gain. Furthermore, we order the instances by their scores using \( score_i \) in ascending order. After that, we begin by categorizing the scores (\( \text{score}_i \)) into two classes: \( S^+ \) for positive instances and \( S^- \) for negative instances. Following this, we select the top \( K^+ \) positive candidates and the top \( K^- \) negative candidates, as described in Section \ref{Determining_Number_of_Instances_to_Flip}. Finally, we implement the flips, updating the labels to \( y'_i = 1 - y_i \) for all \( i \) in the combined set of \( K^+ \) and \( K^- \). In the upcoming sections, we will provide a detailed explanation of each term in Equation (\ref{scorei}).

\subsection{Graph Laplacian}\label{Graph_Laplacian}

The graph Laplacian is a fundamental tool for enforcing smoothness constraints in semi-supervised learning and is widely used in manifold regularization frameworks. Its conceptual foundation is elegant: we construct a graph representation where nodes correspond to data points, with edges reflecting pairwise similarities. Once established, the Laplacian can be used as a regularizer that discourages sharp label variations between neighboring nodes. However, in our GFLC approach, we consider the Laplacian term ($L_i$) \citep{streicher2023graph} over the k-NN graph structure, which is defined as:

\begin{equation}\label{eqGL}
L_i = \sum_{j \in \mathcal{N}(i)} w_{ij} \cdot (y_i - y_j)^2
\end{equation}

\noindent where $\mathcal{N}(i)$ represents the set of k-nearest neighbors of instance $x_i$, while $w_{ij}$ denotes the edge weight between nodes $i$ and $j$. The term $(y_i - y_j)^2$ penalizes label disagreement between connected nodes based on their labels. The quantity $L_i$ calculates the weighted sum of squared differences between the label of node $i$ and those of its neighbors. This serves as a measure of the disagreement between the label of node $i$ and its neighbors, with the weights reflecting the strength of the connections (edge weights). Additionally, the weights $w_{ij}$ in Equation (\ref{eqGL}) are updated according to the update rule defined in Equation (\ref{weight_update_rule4}).

\paragraph{Laplacian Term Analysis:} Higher $L_i$ indicates higher disagreement with neighbors and higher correction priority. In particular, $L_i = 0$ when $y_i = y_j$ for all neighbors $j$ which means perfect local consistency and $L_i$ is maximized when instance $x_i$ has different labels from all its neighbors. Moreover, based on the graph smoothness assumption, nearby instances should have similar labels, high $L_i$ suggests $y_i$ violates local structure and may be incorrect, and correcting such instances improves overall graph consistency.

\paragraph{Majority and Minority Group Nodes:}

In graph-based learning, \textit{majority group nodes} refer to densely connected clusters where nodes share similar characteristics (e.g., a predominant class in classification tasks). Conversely, \textit{minority group nodes} represent sparsely connected regions with distinctive properties (e.g., under-represented classes). The majority group nodes typically have higher degree connections within their group (intra-group edges). Conversely, minority nodes often exhibit more connections \textit{across} groups (inter-group edges) (i.e.\ more edges to neighbors whose values \(y_j\) differ). Hence, $L_i$ will typically be larger for those minority nodes than for the majority nodes. In other words, we are considering the following situation: (i) Majority nodes usually sit in a tight cluster where most neighbors \(j\) also belong to the same class (or have very similar \(y_j\)). That means \((y_i - y_j)^2\) is small on almost every intra‐group edge, so their total \(\sum w_{ij}(y_i - y_j)^2\) stays relatively low. (ii) Minority nodes, by contrast, often have fewer same‐group neighbors and relatively more edges “across” to the majority. Along each of those inter‐group edges, \(y_i\) and \(y_j\) are more likely to differ, so \((y_i - y_j)^2\) is larger. Even if the weights \(w_{ij}\) are the same size, having more of those “large‐difference” terms in the sum drives \(L_i\) up. Therefore, nodes with higher \(L_i\) get penalized more heavily. A minority node’s tendency to connect across groups thus “costs” it more in the Laplacian penalty than a majority node that mostly connects inside its own (homogeneous) cluster.\\

Next, we explain the following cases for the inter-group edges and their connection to our GFLC approach:

\paragraph{Inter-group edges:} The weight reduction after updating on inter-group edges ($y_i \neq y_j$) when Forman curvature $F(e_{ij}) < 0$, acts as an \textit{adaptive smoothing filter}. Thus, this helps for noise robustness by making spurious connections between groups (e.g., mislabeled nodes) experience weight decay.

\paragraph{Strong inter-group edges:} Suppose there exists at least one neighbor \(j\) for which $y_i \neq y_j$ but $w_{ij}$is very large. Since \((y_i - y_j)^2 = 1\) whenever \(y_i\) and \(y_j\) differ, the term $ w_{ij}\,(y_i - y_j)^2 $ becomes large due to the high weight \(w_{ij}\). Consequently, that single edge contributes a large value to the sum, driving \(L_i\) to be high. In other words, a large-weight edge prefers \(y_i = y_j\), so observing \(y_i \neq y_j\) despite \(w_{ij}\) being large is a strong indicator that \(y_i\) itself is likely erroneous. Therefore, in our GFLC approach, high \(L_i\) is a signal that node \(i\) is a candidate for having a mislabeled \(y_i\).

\subsection{Margin Term}

In this section, we define the margin term ($1 - M_i$) which converts prediction confidence into a correction priority score, ensuring that we focus on the most uncertain predictions where label noise is most likely to occur. For that, we train an ensemble model (e.g., Random Forest) to obtain initial predictions, then we optimize the class threshold via ROC analysis. Moreover, we define $M_i$ for each node $i$ based on its predicted probability $p_i\in[0,1]$. where $p_i\in[0,1]$ is the predicted probability, and $[\tau^-, \tau^+]$ defines the ambiguous region. Thus, $M_i$ measures how far $p_i$ lies outside the central ambiguous region $[\tau^-,\tau^+]$. Intuitively, large positive $M_i$ indicates a confident prediction that may override a noisy label.

\begin{equation}
\tau^+ = \underset{\tau}{\text{argmax}} \left(\text{TPR}(\tau) - 5\cdot\text{FPR}(\tau)\right)  \quad  \quad and \quad  \quad \tau^- = 1 - \tau^+
\end{equation}

\begin{equation}
M_i = \begin{cases}
p_i - \tau^+ & p_i \geq \tau^+ \\
\tau^- - p_i & p_i \leq \tau^- \\
0 & \text{otherwise}
\end{cases}
\end{equation}

\noindent Higher margin term ($1 - M_i$) values are better for correction purposes because: larger $\text{(margin term)}_i$ indicates the model is less confident about sample $i$ (high uncertainty), and less confident samples are more likely to have noisy/incorrect labels. Therefore, they should be prioritized for potential label correction.

\subsection{Fairness Term}\label{Fairness_Scoring}

\paragraph{Fairness Term ($\mathrm{DP}_i$).} The fairness term measures the impact of flipping label $y_i$ on demographic parity \citep{dwork2012fairness} across sensitive groups and is defined as follows:

\begin{equation}
\Delta\mathrm{DP}_i= \mathrm{DP}^{\text{new}} - \mathrm{DP}^{\text{original}}
\end{equation}

Both $\mathrm{DP}^{\text{new}}$ and $\mathrm{DP}^{\text{original}}$ measure demographic parity ($\mathrm{DP}$) and are computed as the ratio between the minimum and maximum positive rates based on the sensitive group \citep{e2024fair}. $\mathrm{DP}^{\text{original}}$ is the current demographic parity ratio and $\mathrm{DP}^{\text{new}}$ is the hypothetical demographic parity ratio after flipping $y_i$. Both ratios range from 0 to 1, where 1 indicates perfect demographic parity. Furthermore, higher values for $\mathrm{DP}$ indicate more balanced positive rates across groups. Mathematically, we define demographic parity as follows:

\begin{equation}\label{DP_equ}
  \mathrm{DP}
  = \frac{\displaystyle \min_s \bigl(\Pr(\hat y = 1 \mid S = s)\bigr)}
         {\displaystyle \max_s \bigl(\Pr(\hat y = 1 \mid S = s)\bigr)}
\end{equation}

The resulting fairness term, \( \Delta \mathrm{DP}_i \), provides the following insights: (i) $\Delta\mathrm{DP}_i > 0$ means that flipping $y_i$ improves demographic parity (moves closer to balance), (ii) $\Delta\mathrm{DP}_i< 0$ means that flipping $y_i$ worsens demographic parity (moves away from balance) and (iii) $\Delta\mathrm{DP}_i = 0$ means that flipping $y_i$ has no impact on demographic parity.

\subsection{Determining Number of Instances to Flip}\label{Determining_Number_of_Instances_to_Flip}

The previous scoring function ensures that among all possible flips, we select the \textit{most appropriate} candidates. This addresses the critical question: \textit{'How many instances should be flipped?'}. In order to determine the total number of positive instances to flip to negative (top $K^-$) and the total number of negative instances to flip to positive (top $K^+$), we consider using some parts of Fair-OBNC \citep{e2024fair} to achieve this as follows:

\paragraph{First,} we define the overall positive rate as \(  P(y) = \frac{1}{n}\sum_{i=1}^{n} y_i \). Moreover, we define the minimum and maximum prevalence values as \( P_{\text{lower}} = P(y) \cdot (1 - D) \) and \( P_{\text{upper}} = P(y) \cdot (1 + D) \), where $D \in [0,1]$ is the disparity tolerance parameter.

\paragraph{Second,} for each sensitive group \( s \in \mathcal{S} \) we define \( n_s = |\{i : s_i = s\}| \) as group size and \(  P(y|s) = \frac{1}{n_s}\sum_{i: s_i = s} y_i \quad \) as group positive rate.

\paragraph{Third,} we use the following flip determination logic \citep{e2024fair}:

\begin{equation}
\text{For each group } s: \quad
\begin{cases}
\text{If } P(y|s) < P_{\text{lower}}: & \text{Flip negatives to positives} \\
\text{If } P(y|s) > P_{\text{upper}}: & \text{Flip positives to negatives} \\
\text{Otherwise}: & \text{No flips needed}
\end{cases}
\end{equation}

\paragraph{Finally,} we are computing the required flips $K^+$ and $K^-$ as follows:

\begin{align}
K^+ &= \sum_{s: P(y|s) < P_{\text{lower}}} \left\lceil n_s \cdot (P_{\text{lower}} - P(y|s)) \right\rceil \\
K^- &= \sum_{s: P(y|s) > P_{\text{upper}}} \left\lceil n_s \cdot (P(y|s) - P_{\text{upper}}) \right\rceil
\end{align}

\section{Experimental setup}\label{Experimental_setup}

In this section, we provide all the experimental setups that we used to examine how our proposed method (GFLC) learns a robust and fair classifier in the presence of label noise.

\subsection{Dataset}

In our experiments, we used the bank account fraud dataset \citep{jesus2022turning}, specifically Variant II of the suite that is designed for binary classification and focuses on detecting fraud in bank account openings. The applicant's age serves as the sensitive attribute, represented as a binary variable. In addition, applicants aged 50 and older constitute the first sensitive group $(s = A)$, while those under 50 belong to the second group $(s = B)$. We used the same data set settings as those described in \citep{e2024fair}. The dataset consists of one million individual application instances, using thirty features \citep{jesus2022turning}. Moreover, the positive class (fraud) comprises approximately 1.15\% of the dataset, while the negative class (legitimate) makes up about 98.85\%.

\subsubsection{Label noise injection.}

To evaluate the effectiveness of our GFLC method compared to the baseline across various biased data scenarios, we inject label noise into unbiased versions of the dataset as done in the works \citep{silva2023systematic, e2024fair}. The process outlined in the work \citep{e2024fair} begins with the generation of an independent and identically distributed (IID) dataset concerning the sensitive attribute and the data splits. First, the sensitive attribute column is shuffled to eliminate any existing relationships between this column, the labels, and the features. Next, the instances are randomly shuffled into training, validation, and test sets to prevent potential data drift in the original splits. The subsequent injection of label noise is based on both the label and the sensitive group to which each instance belongs.

\paragraph{Label Noise.} In our experiments, we focus exclusively on the scenario of applying label noise equally to both labels. This differs from the approach taken in \citep{e2024fair}, where three scenarios were considered: applying label noise equally to both labels, applying noise to samples with positive labels, and applying noise to samples with negative labels. Nevertheless, the study by \citep{e2024fair} found that when only adding noise to the positive label instances, the performance of the models showed little variation across different methods and noise rates and this is likely due to the small number of positive cases (and, therefore, mislabeled instances) in the dataset. Additionally, the results for the other two scenarios—injecting noise into both labels equally and applying it only to samples with negative labels—were very similar. As a result, similar conclusions can be drawn when analyzing the effects of injecting noise solely into the negative label and across both labels \citep{e2024fair}. Furthermore, as noted in \citep{e2024fair}, we examine three different noise rates: 5\%, 10\%, and 20\%.

\paragraph{Sensitive group.} In the context of sensitive groups, as discussed in \citep{e2024fair}, we apply noise uniformly to instances in one specific group (Group A only), simulating a scenario where the sensitive attribute influences the likelihood of receiving an incorrect label.

\subsubsection{IID Test Set.}\label{IID_Test_Set0} In the domain of learning with noisy labels, it is essential to evaluate the trained models using a clean IID test set after obtaining corrected labels for a noisy training dataset and training models on this corrected dataset. This evaluation helps us understand how the models perform in the case where the biases present in the training data are mitigated in the new resulting train dataset \citep{e2024fair}.

\subsection{GFLC}

In our GFLC method, we are using the following parameters in all experimental settings. For K-NN graph (Section \ref{k-NN_graph}) we are using $k=10$. Moreover, for Equation (\ref{scorei}) we are using $\alpha=0.2$, $\beta=0.6$ and $\gamma=0.2$. In Equation (\ref{weight_update_rule4}) the weights are updated for 2 iterations ($ricci\_iter=2$). Furthermore, we select $D=0.05$ (Section \ref{Determining_Number_of_Instances_to_Flip}) as the target of disparity. Finally, after obtaining the modified/corrected training data set using our proposed GFLC method (Section \ref{section:GFLC}), we train a LightGBM model configured with 100 estimators and a learning rate of 0.1 using the corrected training data set. The final trained model will be used to make predictions for the IID test set (Section \ref{IID_Test_Set0}).

\subsection{Baseline: Fair-OBNC}\label{Fair_OBNC_sec} 
We use Fair Ordering-Based Noise Correction (Fair-OBNC) \citep{e2024fair} as a baseline. It is fairness-aware label noise correction method, extending the OBNC algorithm \citep{feng2015class} with fairness considerations. This method is used to enhance demographic parity in training datasets, making it a good baseline for the intersection of fairness and noise correction. For Fair-OBNC, we follow the same experimental settings outlined in \citep{e2024fair}, where 50 hyperparameter configurations are randomly sampled and applied to the training set. However, unlike \citep{e2024fair}, which used a fixed decision threshold for this process, we utilize the same procedure for various decision thresholds ranging from 0 to 0.6 in our experiments. Furthermore, after obtaining the modified training sets, LightGBM models are used to be be trained on them \citep{e2024fair}. These models then make predictions for the IID clean test set, and finally, we average the performance and fairness metrics across the 50 runs. Additionally, the implementation code for Fair-OBNC is available in the Aequitas package\footnote{\url{https://github.com/dssg/aequitas}} under the name LabelFlipping.

\subsection{Metrics}

In this section, we outline the performance and fairness metrics used in our experiments to evaluate GFLC and Fair-OBNC approaches on the Bank Account Fraud dataset. The dataset focuses on detecting fraud during the account opening process, with the primary objective of identifying as many fraudulent attempts as possible \citep{e2024fair}. 

\subsubsection{Performance Metrics}\label{sec:performance_metrics1}

\paragraph{The Area Under the Curve (AUC):} The AUC metriccan be used to measure the area under the Receiver Operating Characteristic (ROC) curve. In general, the ROC plots the True Positive Rate against the False Positive Rate at various threshold settings. Furthermore, AUC score ranges from 0 to 1 where a score of 0.5 indicates random guessing, while a score of 1 means perfect performance.

\paragraph{True Positive Rate (TPR):} TPR, which is also known as sensitivity or recall, is a key performance metric in binary classification. It measures the proportion of actual positive cases that the model correctly identifies. Mathematically, TPR is defined as: TPR = TP / (TP + FN). In this formula, TP represents true positives, and FN represents false negatives. Essentially, TPR answers the question: out of all the actual positive cases in the dataset, how many did the model successfully identify?

\paragraph{True Negative Rate (TNR):}

The True Negative Rate (TNR), also known as specificity, measures the proportion of actual negative cases that are correctly identified as negative by the model. Mathematically expressed as TNR = TN / (TN + FP), where TN represents true negatives and FP represents false positives. Therefore, the TNR addresses the question: Of all the actual negative cases in the dataset, how many did the model correctly classify as negative?

\paragraph{False Positive Rate (FPR):} The False Positive Rate (FPR) measures the proportion of actual negative cases incorrectly identified as positive by a model . FPR can be calculated as the ratio of false positives (FP) to the sum of false positives and true negatives (TN), expressed as FPR=FP/(FP+TN).

\paragraph{False Negative Rate (FNR):} The False Negative Rate (FNR) represents the proportion of actual positive cases erroneously classified as negative. It is computed as the ratio of false negatives (FN) to the sum of false negatives and true positives (TP), given by 
FNR=FN/(FN+TP).

\paragraph{Precision.} Precision measures the proportion of true positive predictions among all positive predictions: $\frac{TP}{TP+FP}$. It is particularly important in applications where false positives are costly.

\subsubsection{Fairness Metrics.}\label{sec:performance_metrics22}

Since our method (GFLC) and the baseline (Fair-OBNC) \citep{e2024fair} (Sections \ref{Fairness_Scoring} and \ref{Fair_OBNC_sec}) primarily focus on achieving \textit{Demographic Parity} as the key fairness criterion during the label correction process, we utilized the Demographic Parity metric to evaluate the group fairness of our approach on the dataset. However, we also considered \textit{Equal Opportunity} and \textit{Equalized Odds} metrics, even though they were not the main focus of our experiments. By examining these metrics, we aim to provide a thorough evaluation of fairness in our models, ensuring they treat all groups equitably.

\paragraph{Demographic Parity.} To compute the Demographic Parity metric (Equation \ref{eq:demographic_parity}), we calculate the ratio between the lowest and highest predicted prevalence (Equation \ref{DP_equ}). In this metric, values close to 1 represent an equilibrium of predicted prevalence between every group in the dataset, while values close to 0 represent higher disparities in the fairness metric. In our experiments, this metric is called $pprev\_ratio$ (predicted prevalence) same as it is named in the Aequitas toolkit \citep{saleiro2018aequitas, jesus2024aequitas, e2024fair}.

\begin{equation}
P(\hat{y} = 1|s = A) = P(\hat{y} = 1|s = B)
\label{eq:demographic_parity}
\end{equation}

\paragraph{Equal Opportunity.} Equal Opportunity \citep{hardt2016equality} requires that a model achieves the same true positive rate (TPR) for different subgroups when considering only instances with a positive label \citep{hardt2016equality}. Formally, it is defined as:
\[
P(\hat{Y} = 1 | S = 0, Y = 1) = P(\hat{Y} = 1 | S = 1, Y = 1),
\]
where \( \hat{Y} \) represents the predicted outcome, \( S \) is the sensitive attribute, and \( Y \) is the true label. This condition ensures that individuals from different groups who are actually positive (i.e., have a positive true label) have an equal probability of being classified as positive by the model.

\paragraph{Equalized Odds.}
It extends the concept of Equal Opportunity by requiring that both the true positive rate (TPR) and the false positive rate (FPR) be equal across different groups \citep{hardt2016equality}. It can be expressed as:
\[
P(\hat{Y} = y | S = 0, Y = y) = P(\hat{Y} = y | S = 1, Y = y), \quad y \in \{0, 1\}.
\]
The Equalized Odds metric ensures that the model's performance is consistent across groups in terms of both correctly identifying positives and avoiding false positives.

\section{Results}\label{Results_sec} In the following subsections, we present and explain the results of our proposed GFLC approach in comparison to the baseline method, Fair-OBNC. From each approach, we obtain a trained model that was trained on a training dataset, which includes the corresponding corrected training labels. Moreover, we then test the final trained models from each approach on the same IID test set.

\subsection{Evaluation of performance}
Table~\ref{tab:auc_noise_comparison34567} reports the AUC scores of GFLC and Fair‑OBNC under varying noise levels. At 5\% noise, GFLC achieves an AUC of 0.874 versus 0.813 for Fair‑OBNC. When the noise rate rises to 10\%, GFLC maintains its lead with an AUC of 0.843 compared to Fair‑OBNC's 0.783. Even at 20\% noise, GFLC outperforms Fair‑OBNC, recording 0.799 against 0.752.

\begin{table}[H] 
\centering
\caption{Performance Comparison Across Different Noise Rates}
\label{tab:auc_noise_comparison34567}
\setlength{\tabcolsep}{12pt}
\begin{tabular}{@{}l*{6}{c}@{}}
\toprule
\multirow{2}{*}{Metric} & \multicolumn{2}{c}{Noise Rate: 5\%} & \multicolumn{2}{c}{Noise Rate: 10\%} & \multicolumn{2}{c}{Noise Rate: 20\%} \\
\cmidrule(lr){2-3} \cmidrule(lr){4-5} \cmidrule(lr){6-7}
& GFLC & Fair-OBNC & GFLC & Fair-OBNC & GFLC & Fair-OBNC \\
\midrule
AUC $\uparrow$ & 0.874 & 0.813 & 0.843 & 0.783 & 0.799 & 0.752 \\
\bottomrule
\end{tabular}
\end{table}

\subsection{Results at Noise Rate 5\%}
\begin{figure}[htbp]
\centering
\begin{subfigure}[b]{0.32\textwidth}
    \centering
    \includegraphics[width=0.95\textwidth]{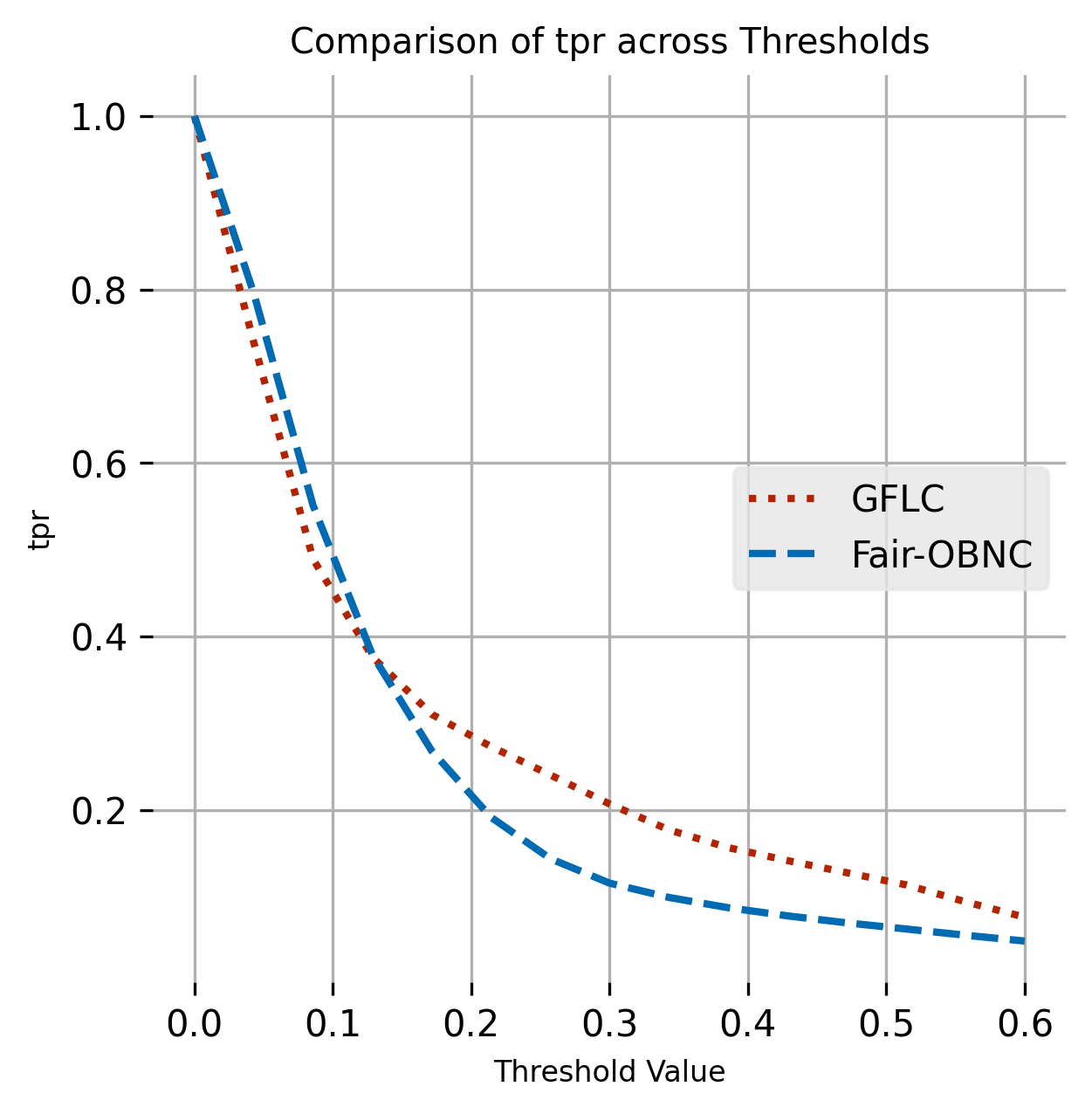}
    \caption{True Positive Rate (TPR)}
    \label{fig:tpr_noise_5}
\end{subfigure}
\hfill
\begin{subfigure}[b]{0.32\textwidth}
    \centering
    \includegraphics[width=0.95\textwidth]{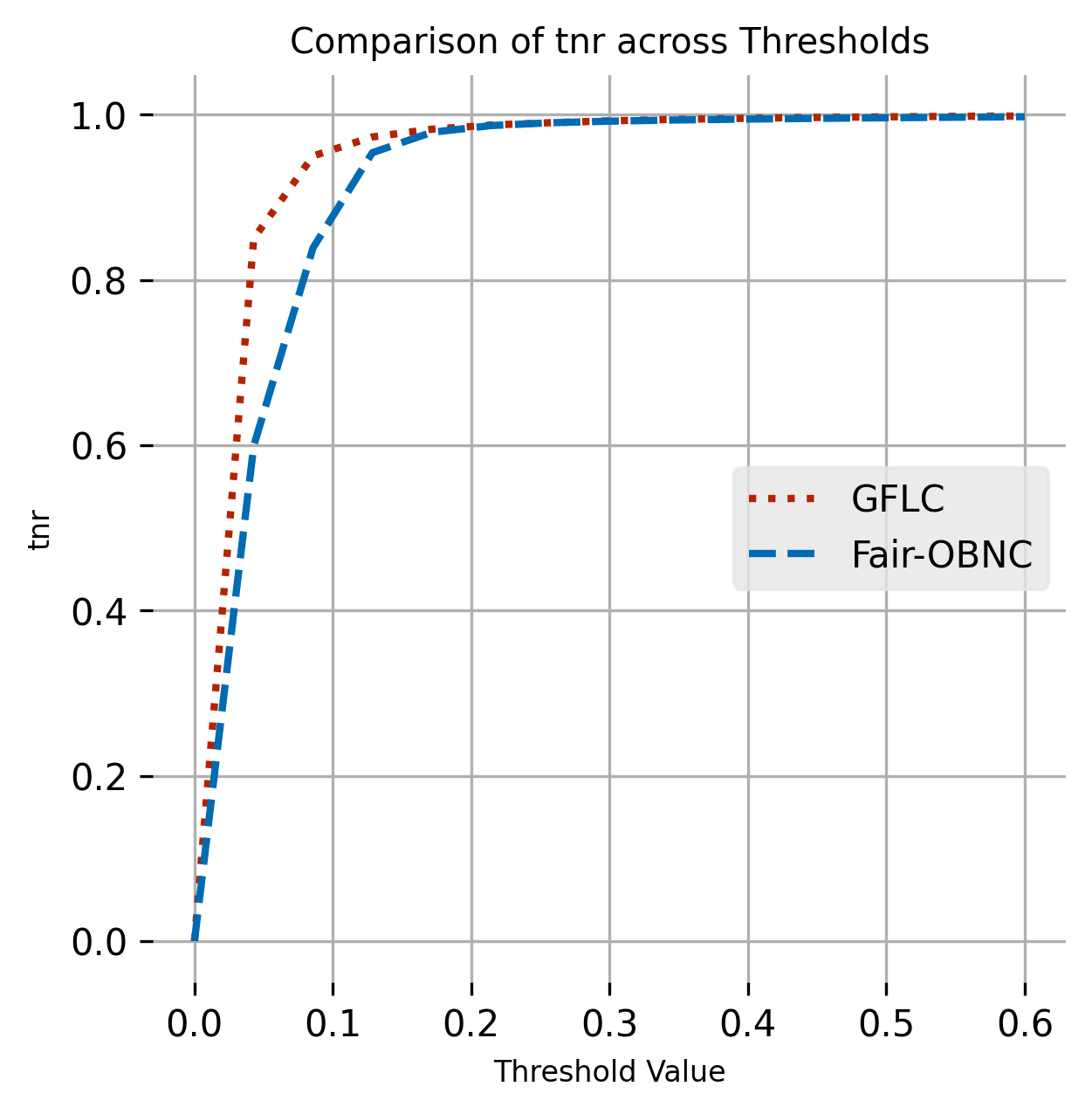}
    \caption{True Negative Rate (TNR)}
    \label{fig:tnr_noise_5}
\end{subfigure}
\hfill
\begin{subfigure}[b]{0.32\textwidth}
    \centering
    \includegraphics[width=0.95\textwidth]{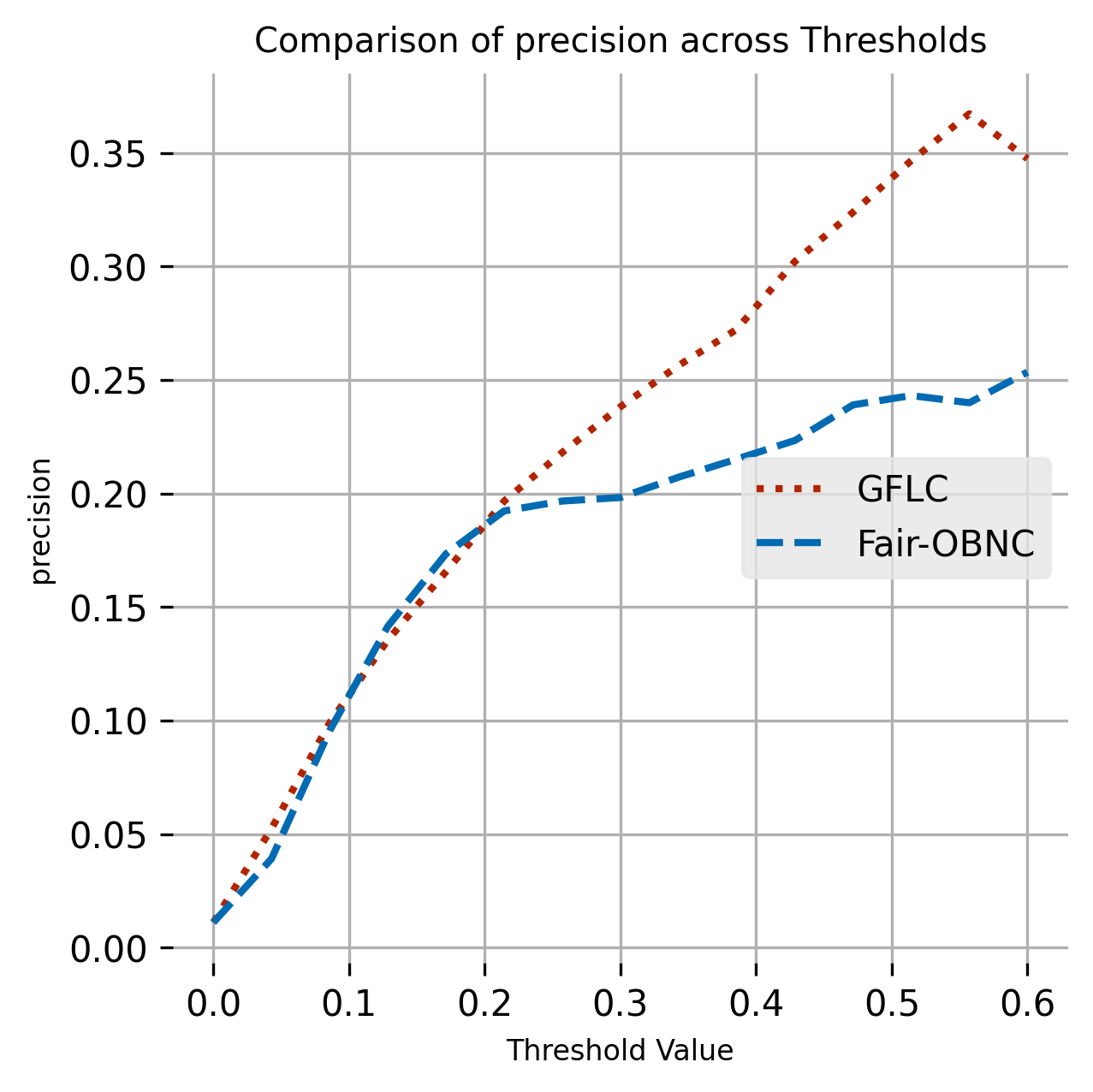}
    \caption{Precision}
    \label{fig:precision_noise_5}
\end{subfigure}

\vspace{0.2cm}

\begin{subfigure}[b]{0.32\textwidth}
    \centering
    \includegraphics[width=0.95\textwidth]{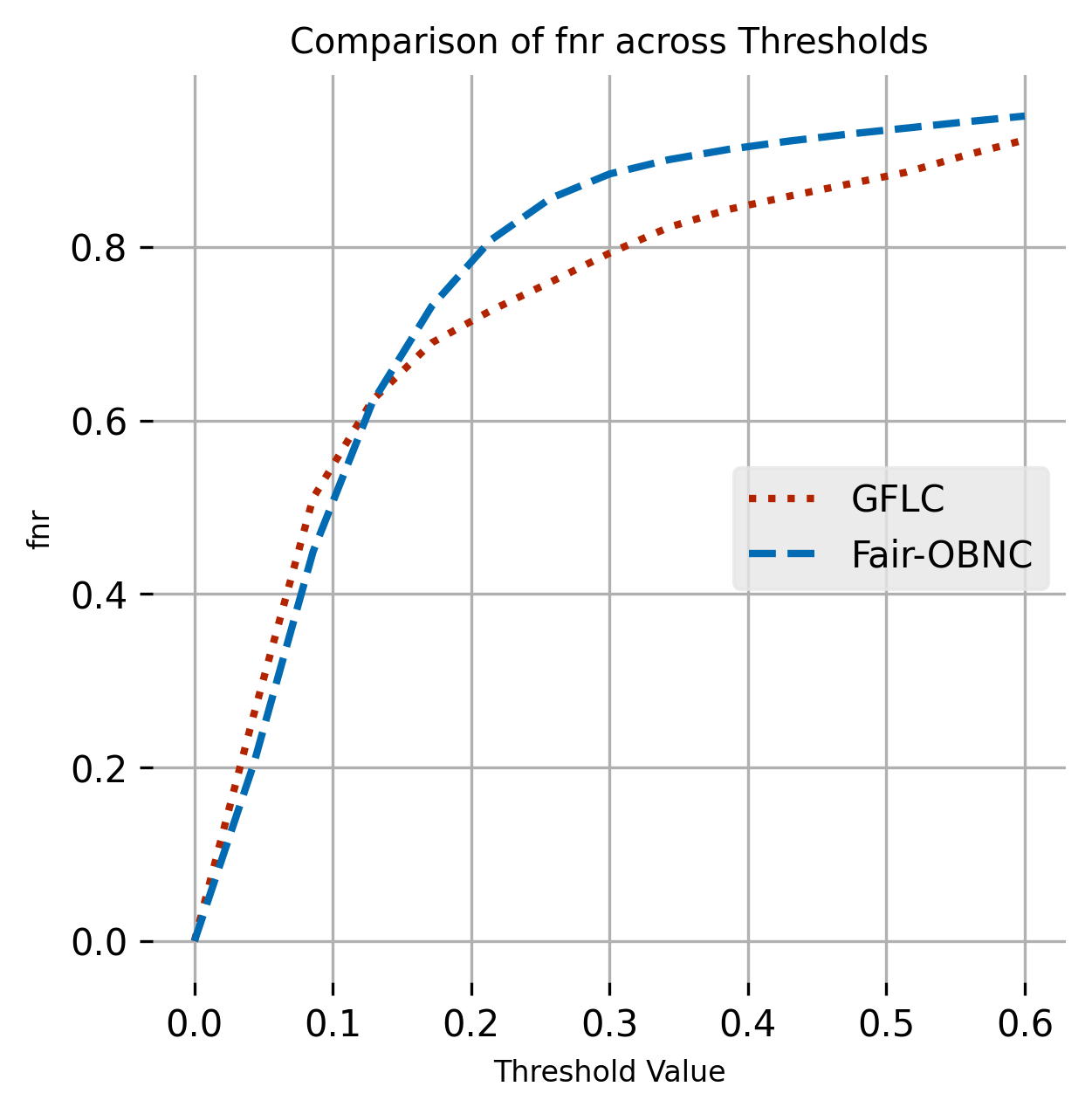}
    \caption{False Negative Rate (FNR)}
    \label{fig:fnr_noise_5}
\end{subfigure}
\hfill
\begin{subfigure}[b]{0.32\textwidth}
    \centering
    \includegraphics[width=0.95\textwidth]{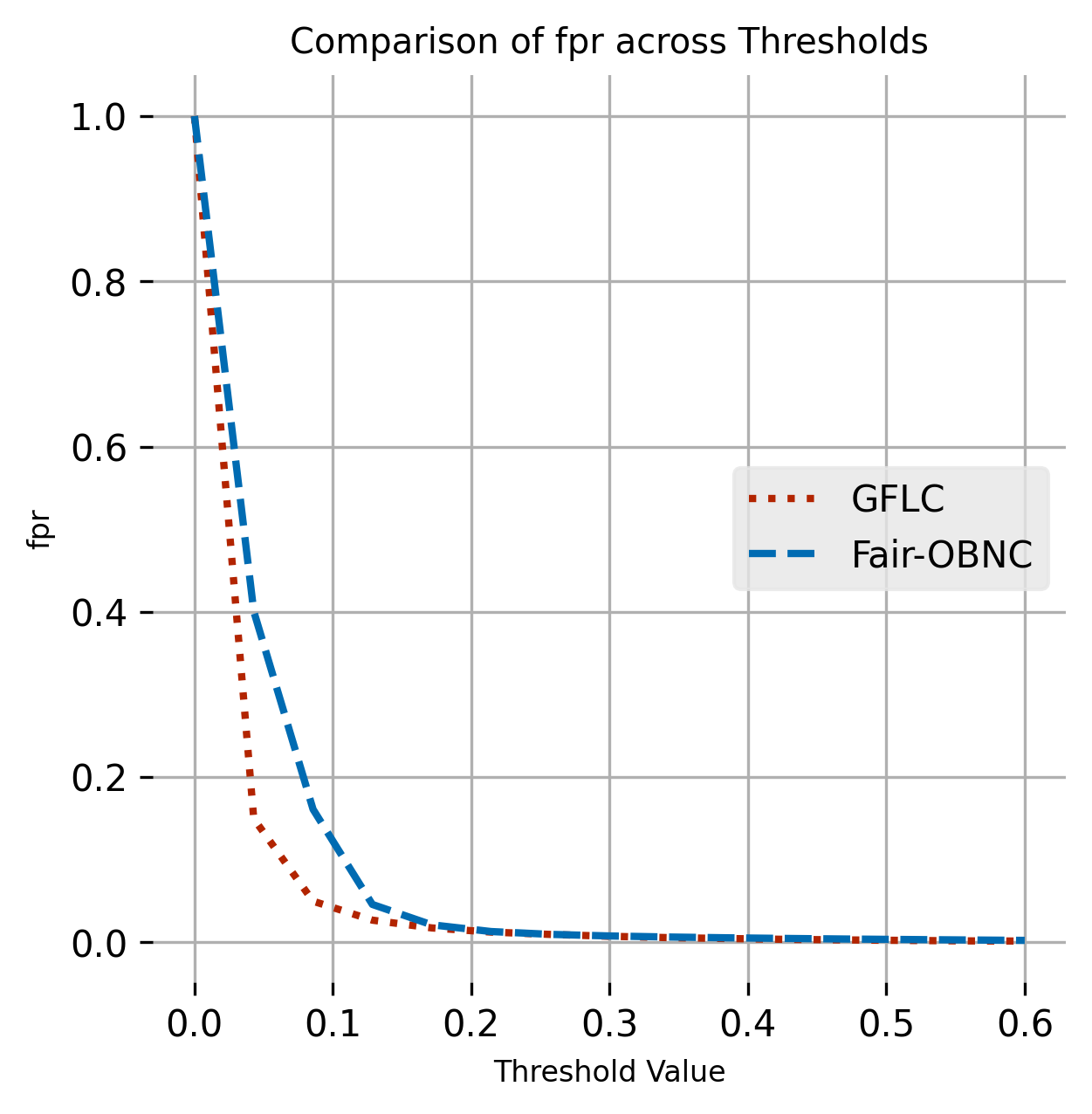}
    \caption{False Positive Rate (FPR)}
    \label{fig:fpr_noise_5}
\end{subfigure}

\caption{Performance Metrics Comparison Across Different Thresholds at 5\% Noise Rate}
\label{fig:performance_metrics_noise_5}
\end{figure}

\begin{figure}[htbp]
\centering
\begin{subfigure}[b]{0.32\textwidth}
    \centering
    \includegraphics[width=0.95\textwidth]{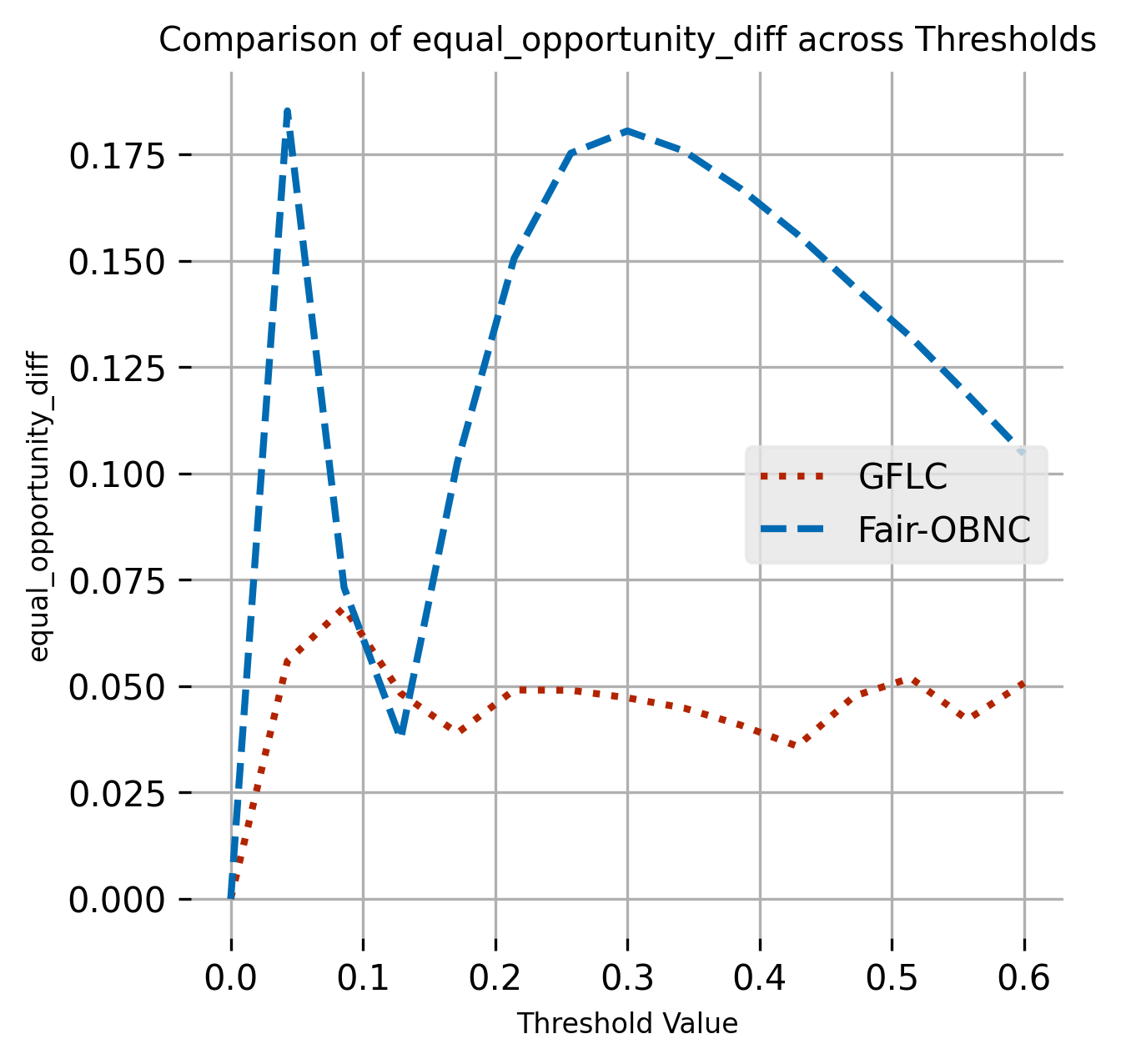}
    \caption{Equal Opportunity $\downarrow$}
    \label{fig:equal_opportunity_noise_5}
\end{subfigure}
\hfill
\begin{subfigure}[b]{0.32\textwidth}
    \centering
    \includegraphics[width=0.95\textwidth]{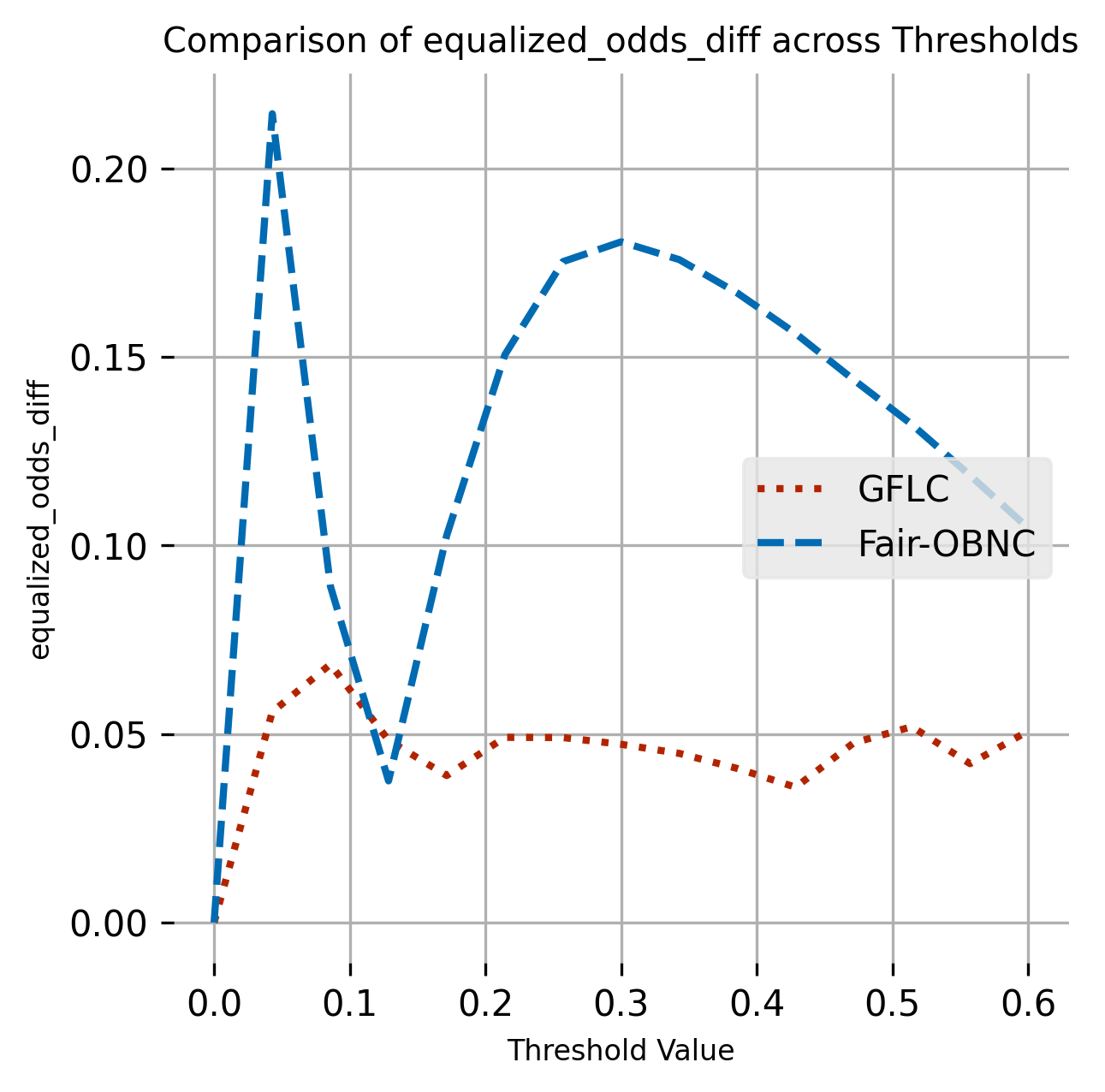}
    \caption{Equalized odds $\downarrow$}
    \label{fig:equalized_odds_noise_5}
\end{subfigure}
\hfill
\begin{subfigure}[b]{0.32\textwidth}
    \centering
    \includegraphics[width=0.95\textwidth]{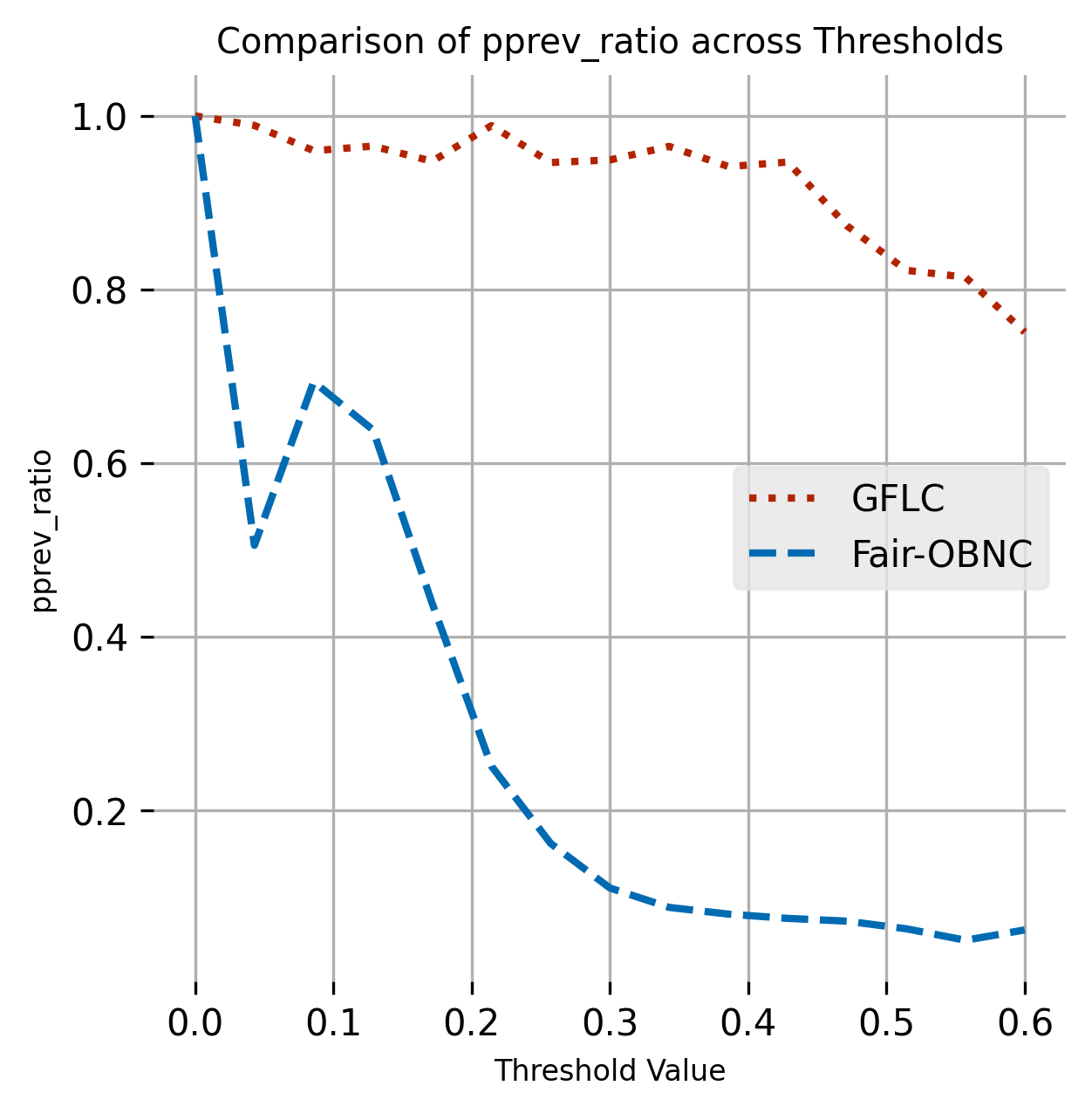}
    \caption{Demographic Parity $\uparrow$}
    \label{fig:pprev_ratio_noise_5}
\end{subfigure}

\caption{Fairness Metrics Comparison Across Different Thresholds at 5\% Noise Rate}
\label{fig:fair_metrics_noise_5}
\end{figure}

The experimental results presented in Figures~(\ref{fig:performance_metrics_noise_5}, ~\ref{fig:fair_metrics_noise_5}) demonstrate a comprehensive comparison between GFLC (red, dotted) and Fair-OBNC (blue, dashed) methods across various fairness and performance metrics at a 5\% label-noise rate. The analysis reveals distinct behavioral patterns and trade-offs between the two approaches across different threshold values in terms of fairness and performance.

\paragraph{True‑Positive Rate (TPR) (Figure~\ref{fig:tpr_noise_5}):}
As the threshold increases, Fair‑OBNC's TPR initially stays slightly higher, down to $\approx$ 0.50 at threshold 0.1 versus GFLC's $\approx$ 0.45. However, after that it drops off faster. Beyond $\approx$ 0.12 threshold, GFLC retains a higher TPR, for instance $\approx$ 0.3 TPR at threshold 0.2 compared to Fair‑OBNC's $\approx$ 0.21.

\paragraph{False‑Positive Rate (FPR) (Figure~\ref{fig:fpr_noise_5}):}
At very low thresholds (near 0) both methods produce nearly 100\% false‑positives (i.e.\ everyone is predicted positive). As the threshold increases, GFLC's FPR drops more sharply: by threshold $\approx 0.1$ it’s already down near 5\%, while Fair‑OBNC is still around 10\%. Beyond $\approx 0.2$, both methods drive FPR very close to zero. This indicates that GFLC results in fewer false alarms for every threshold choice while maintaining a comparable True Positive Rate (TPR) when compared to Fair-OBNC.

\paragraph{Precision (Figure~\ref{fig:precision_noise_5}):}
Low thresholds ($< 0.1$) lead to extremely low precision for both models (because nearly everyone is called positive). In the mid‑range (0.1–0.3), GFLC and Fair‑OBNC track closely until about 0.2, where they both reach $\approx$0.20 precision. At higher thresholds ($ > 0.2$), GFLC pulls ahead—reaching up to $\approx$0.36 precision at threshold 0.55 versus around 0.24 for Fair‑OBNC.

\paragraph{False Negative Rate (FNR):}

Figure~\ref{fig:fnr_noise_5} presents the false negative rate (FNR) across thresholds, where both methods show similar trends but with notable differences in magnitude. Both GFLC and Fair-OBNC demonstrate increasing FNR with higher thresholds. The convergence occurs around threshold 0.15, after which Fair-OBNC maintains higher FNR values.

\paragraph{True Negative Rate (TNR):}

Figure~\ref{fig:tnr_noise_5} demonstrates true negative rate (TNR) performance, where both methods achieve high performance levels above 0.95 for thresholds greater than 0.2. At lower thresholds ($ < 0.2$), GFLC shows slightly faster convergence to optimal TNR values.

\paragraph{Demographic Parity (Figure~\ref{fig:pprev_ratio_noise_5}):}

The demographic parity ratio results in Figure~\ref{fig:pprev_ratio_noise_5} reveal the most dramatic performance differences between the methods, where higher values indicate better demographic parity. GFLC maintains exceptional demographic parity with ratios consistently near 1.0 across all threshold values, indicating optimal demographic balance across groups. Conversely, Fair-OBNC shows substantial demographic parity degradation with increasing thresholds, dropping from approximately 1.0 at threshold 0.0 to roughly 0.05 at threshold 0.6. This represents a severe failure in maintaining demographic parity, suggesting that Fair-OBNC's focus on other fairness metrics comes at an unacceptable cost to demographic balance.

\paragraph{Equal Opportunity (Figure~\ref{fig:equal_opportunity_noise_5}):}

Figure~\ref{fig:equal_opportunity_noise_5} illustrates the equal opportunity difference across varying threshold values, where lower values indicate better fairness. GFLC demonstrates superior and more consistent fairness with equal opportunity differences maintained around 0.05 across all thresholds, indicating stable fairness performance. In contrast, Fair-OBNC exhibits significantly worse fairness performance, with peaks reaching approximately 0.18 at threshold values around 0.05 and 0.3. Moreover, Fair-OBNC shows good Equal Opportunity around 0.04 approximately at threshold 0.11. Furthermore, Fair-OBNC shows a problematic bimodal distribution with substantial fairness degradation at low thresholds (0.05) and moderate thresholds (0.3), suggesting threshold-dependent instability that compromises equal opportunity fairness.

\paragraph{Equalized Odds (Figure~\ref{fig:equalized_odds_noise_5}):}

The equalized odds difference, shown in Figure~\ref{fig:equalized_odds_noise_5}, follows a similar pattern as Equal Opportunity metric. GFLC again demonstrates superior fairness performance with equalized odds differences remaining consistently low around 0.05, showing minimal variation and maintaining stable fairness across all threshold values. Fair-OBNC exhibits poor fairness performance with peaks reaching 0.21 at very low thresholds and 0.175 at moderate thresholds around 0.3. The method shows dramatic fairness deterioration from threshold 0.0 to 0.05, followed by some recovery (approximately at threshold 0.11) but continued instability, indicating unreliable equalized odds performance.

\subsection{Results at Noise Rate 10\%}

\begin{figure}[htbp]
\centering
\begin{subfigure}[b]{0.32\textwidth}
    \centering
    \includegraphics[width=0.95\textwidth]{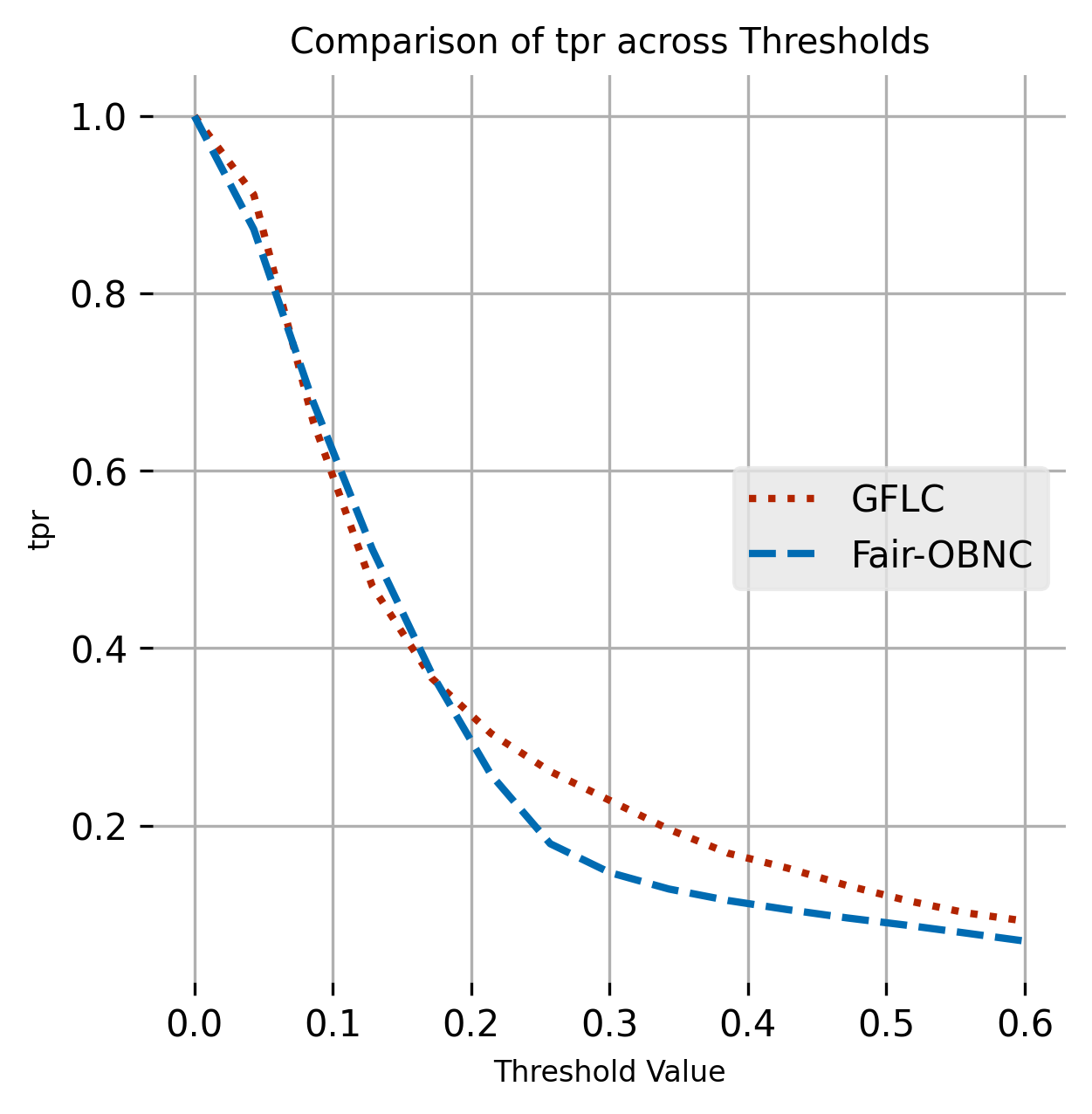}
    \caption{True Positive Rate (TPR)}
    \label{fig:tpr_noise_10}
\end{subfigure}
\hfill
\begin{subfigure}[b]{0.32\textwidth}
    \centering
    \includegraphics[width=0.95\textwidth]{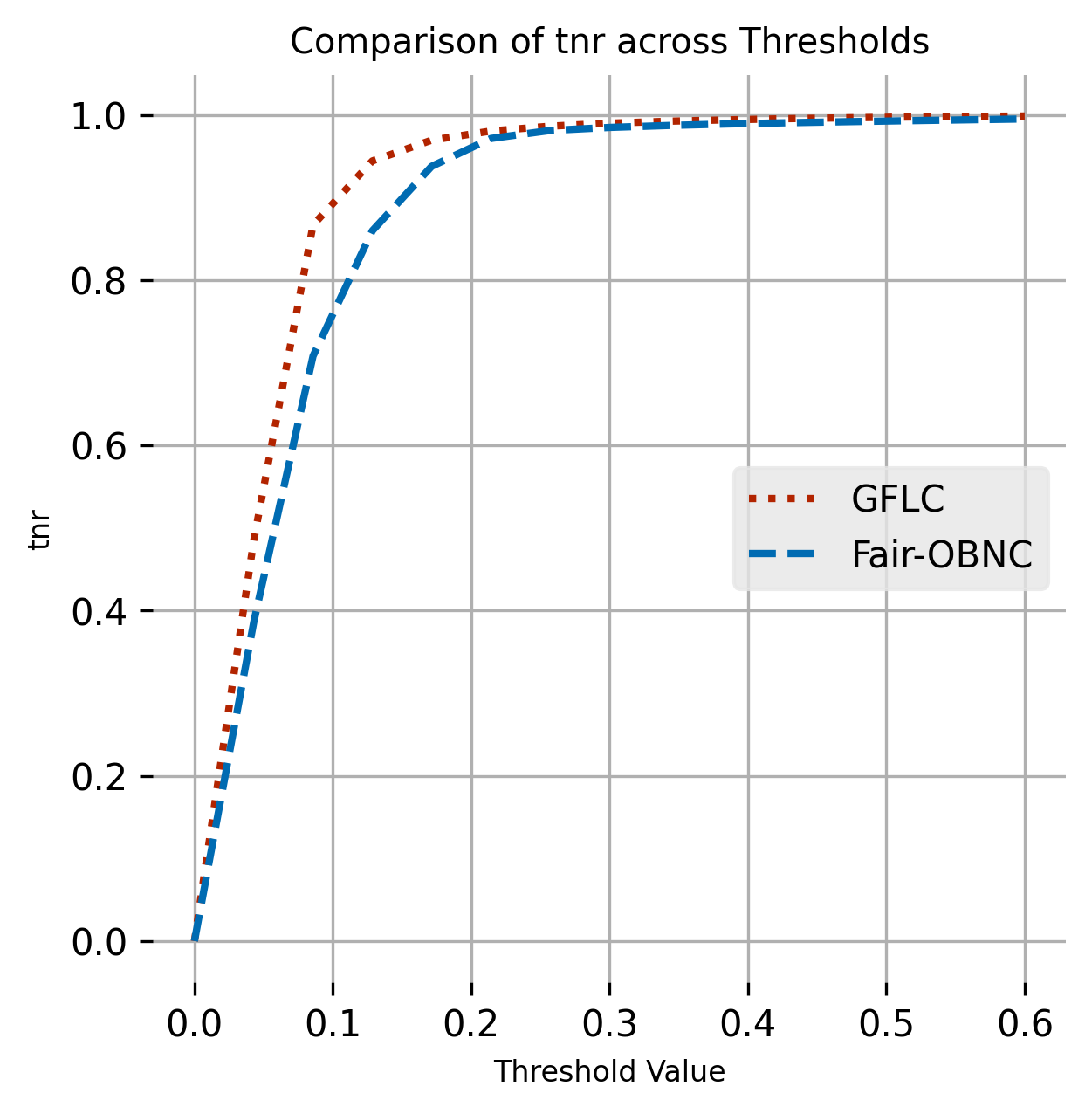}
    \caption{True Negative Rate (TNR)}
    \label{fig:tnr_noise_10}
\end{subfigure}
\hfill
\begin{subfigure}[b]{0.32\textwidth}
    \centering
    \includegraphics[width=0.95\textwidth]{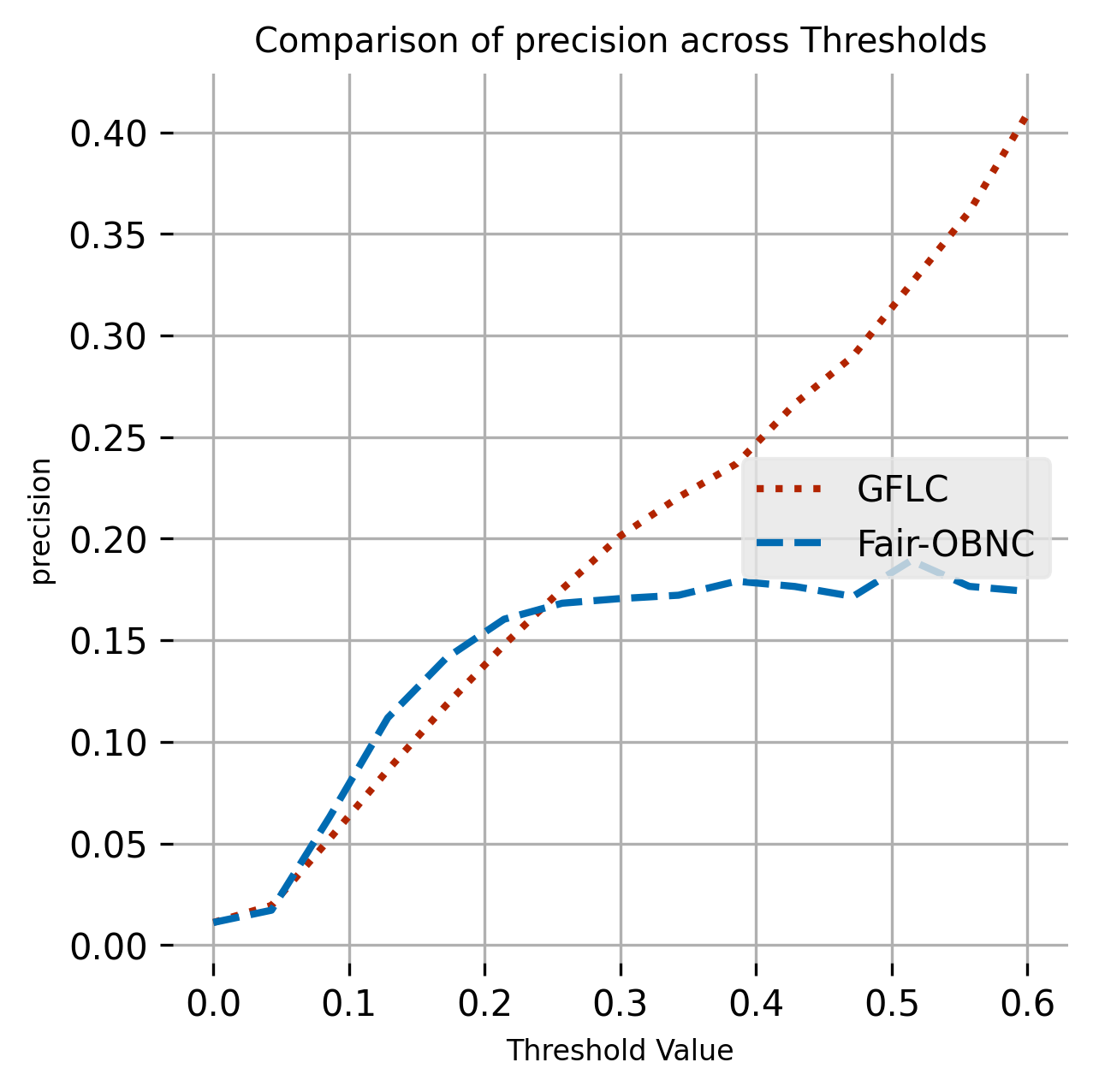}
    \caption{Precision}
    \label{fig:precision_noise_10}
\end{subfigure}

\vspace{0.2cm}

\begin{subfigure}[b]{0.32\textwidth}
    \centering
    \includegraphics[width=0.95\textwidth]{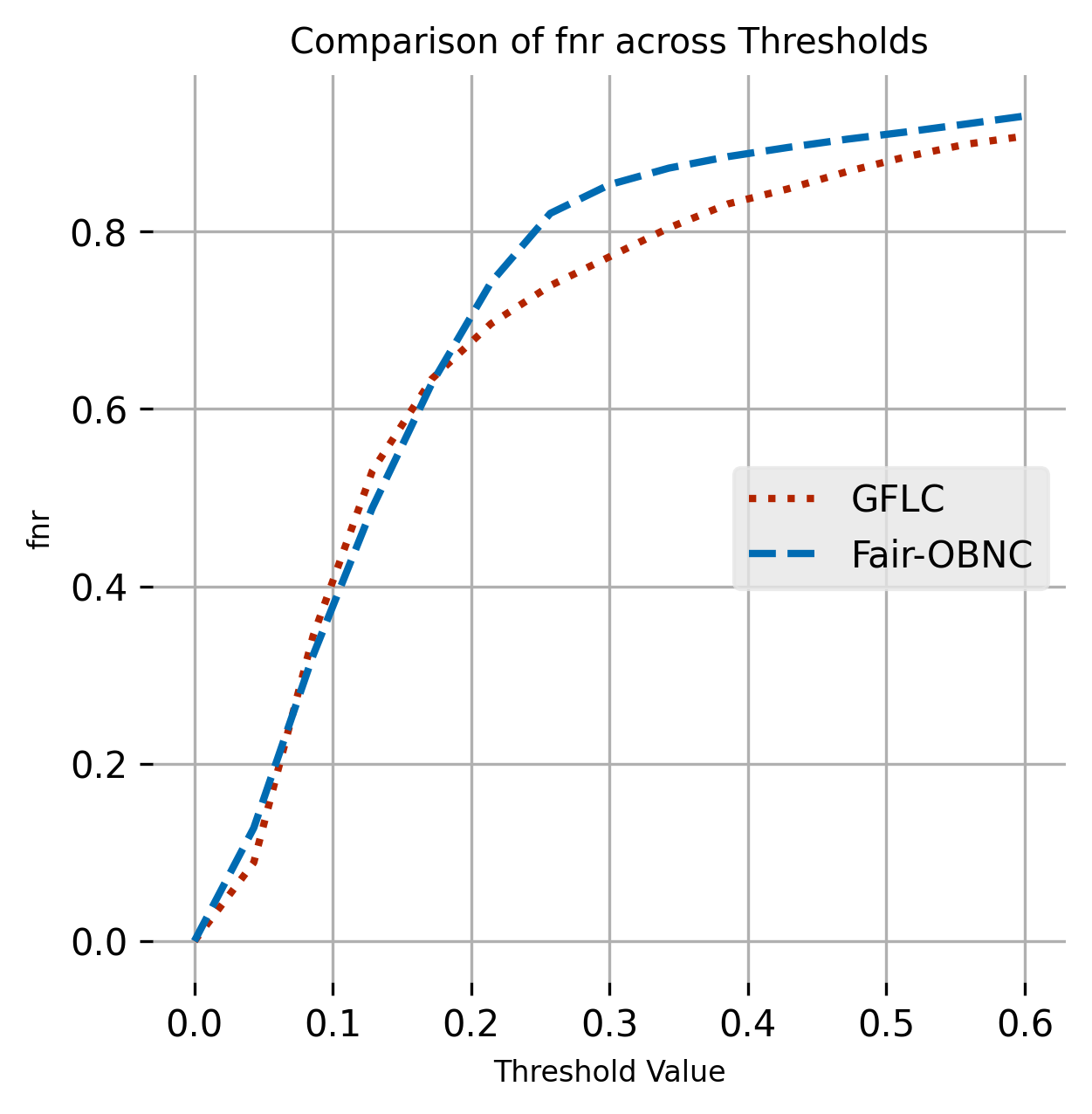}
    \caption{False Negative Rate (FNR)}
    \label{fig:fnr_noise_10}
\end{subfigure}
\hfill
\begin{subfigure}[b]{0.32\textwidth}
    \centering
    \includegraphics[width=0.95\textwidth]{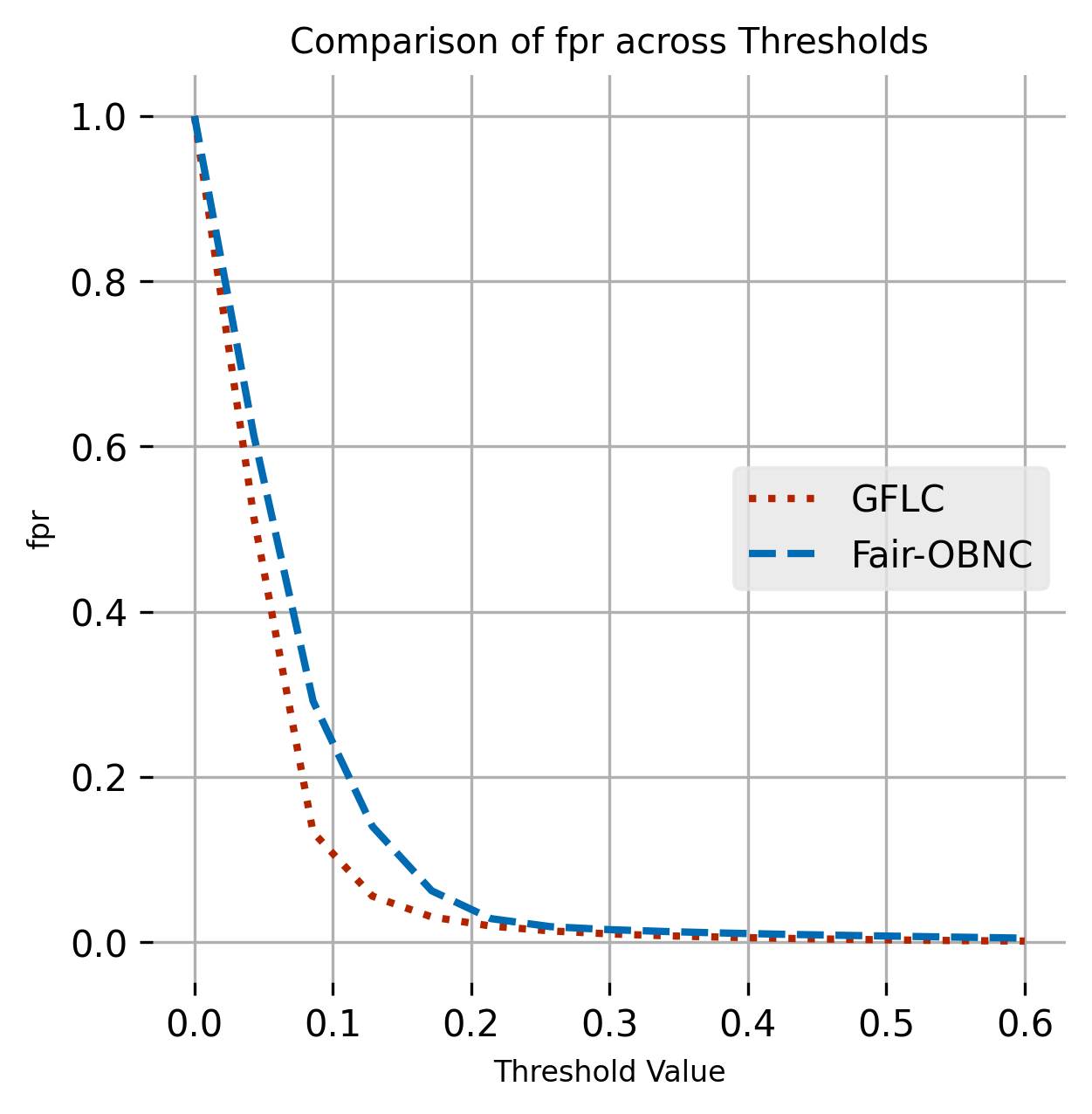}
    \caption{False Positive Rate (FPR)}
    \label{fig:fpr_noise_10}
\end{subfigure}

\caption{Performance Metrics Comparison Across Different Thresholds at 10\% Noise Rate}
\label{fig:performance_metrics_noise_10}
\end{figure}

\begin{figure}[htbp]
\centering
\begin{subfigure}[b]{0.32\textwidth}
    \centering
    \includegraphics[width=0.95\textwidth]{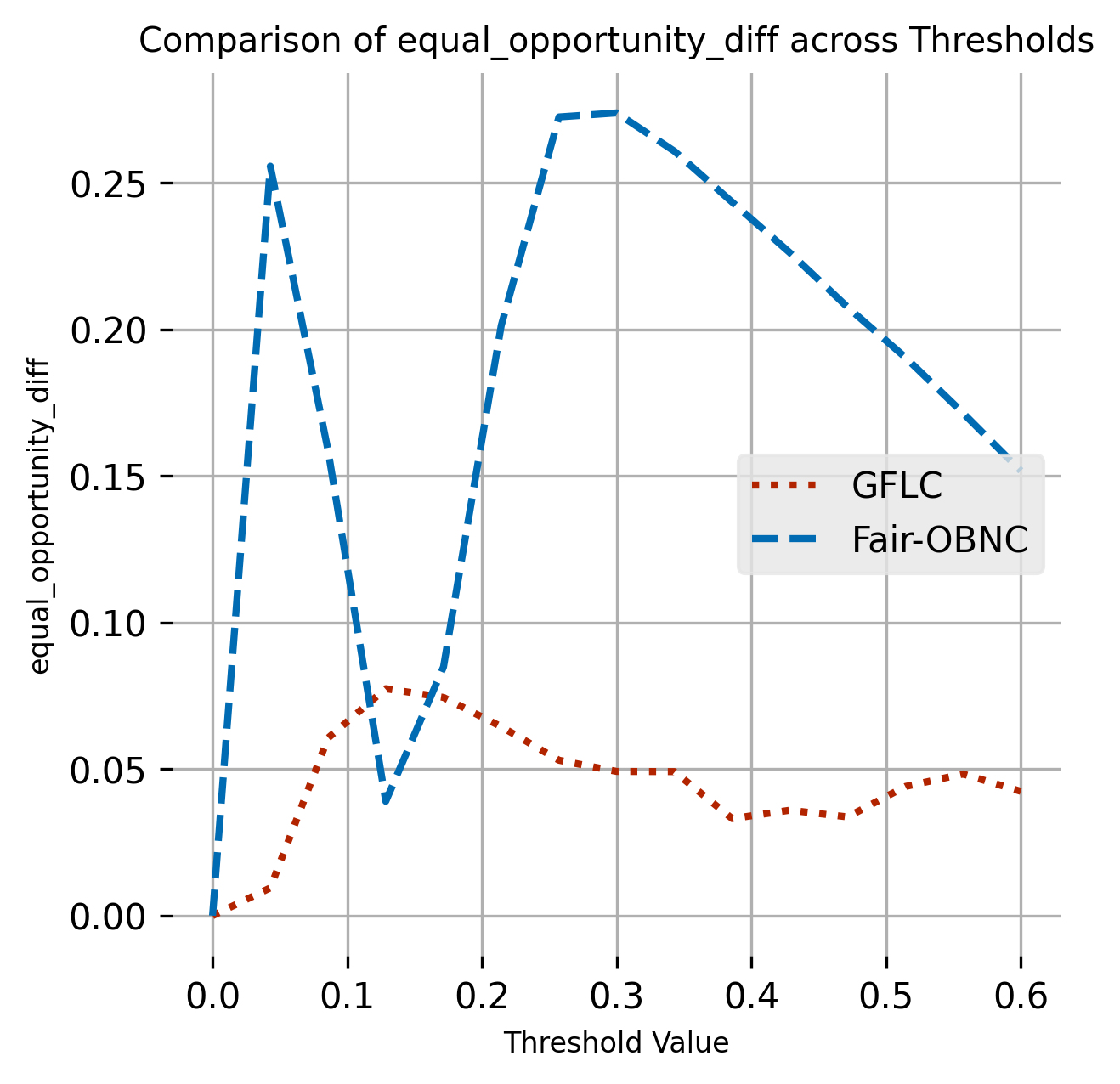}
    \caption{Equal Opportunity $\downarrow$}
    \label{fig:equal_opportunity_noise_10}
\end{subfigure}
\hfill
\begin{subfigure}[b]{0.32\textwidth}
    \centering
    \includegraphics[width=0.95\textwidth]{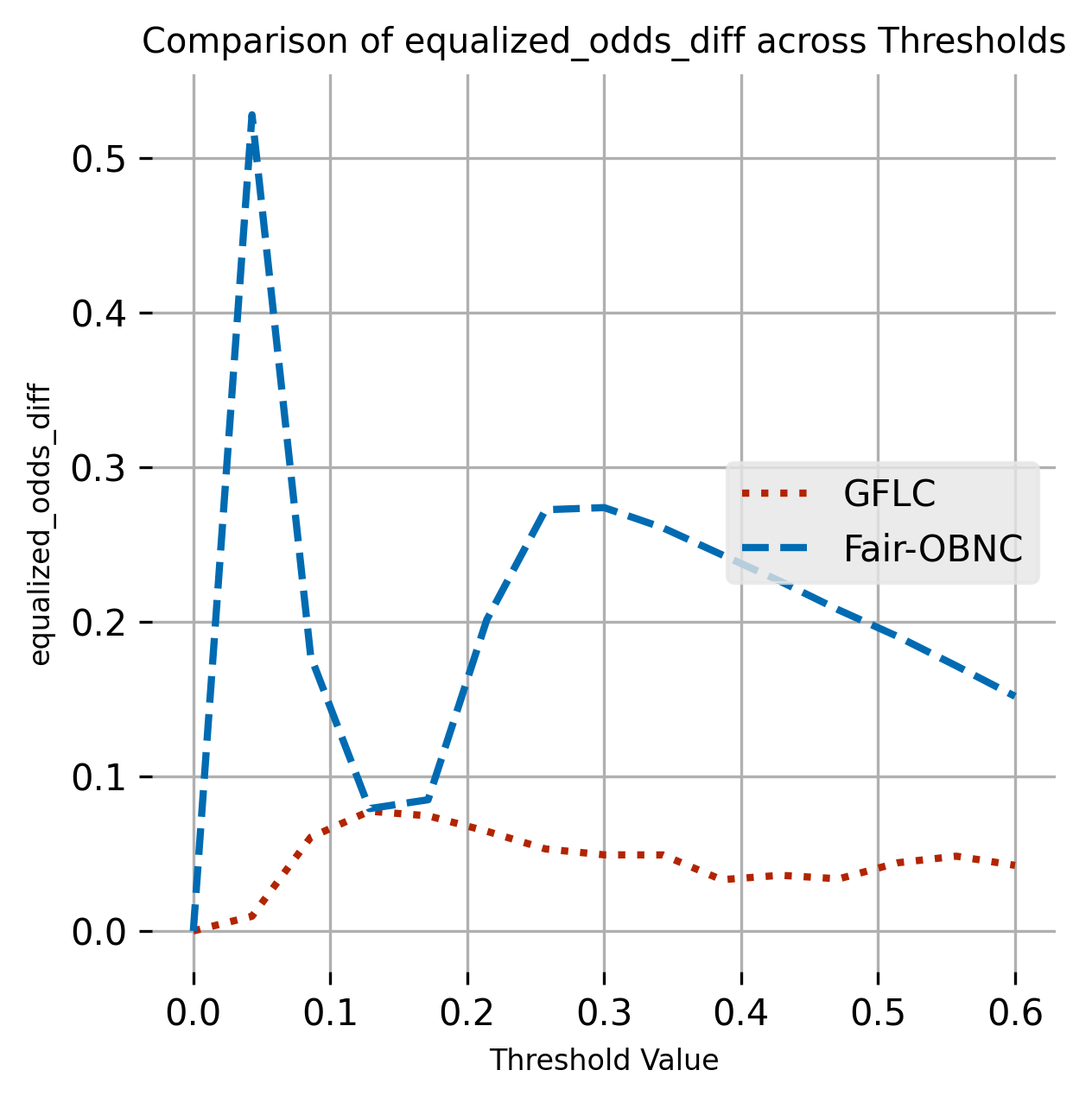}
    \caption{Equalized odds $\downarrow$}
    \label{fig:equalized_odds_noise_10}
\end{subfigure}
\hfill
\begin{subfigure}[b]{0.32\textwidth}
    \centering
    \includegraphics[width=0.95\textwidth]{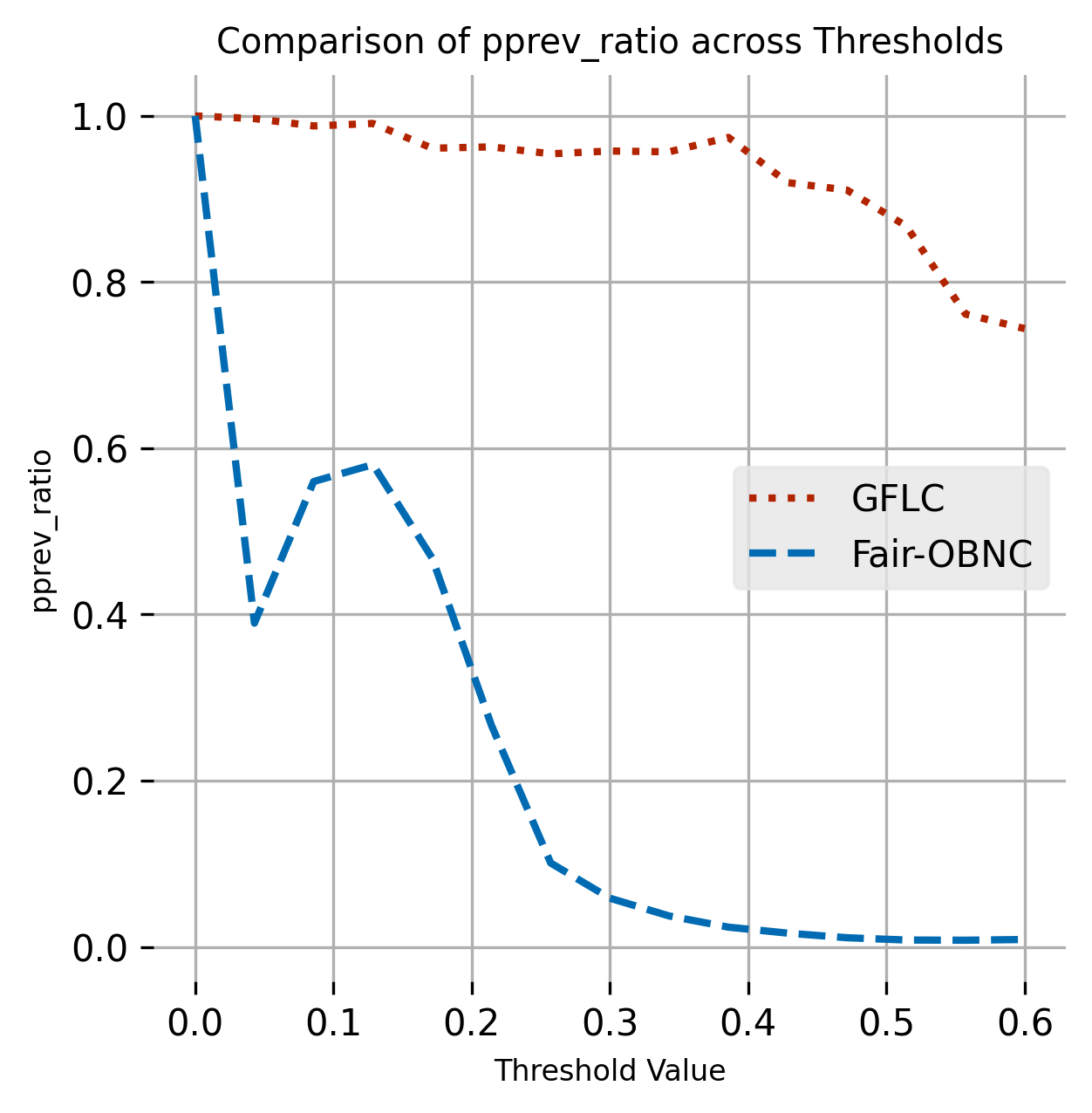}
    \caption{Demographic Parity $\uparrow$}
    \label{fig:pprev_ratio_noise_10}
\end{subfigure}

\caption{Fairness Metrics Comparison Across Different Thresholds at 10\% Noise Rate}
\label{fig:fair_metrics_noise_10}
\end{figure}

At a 10\% label noise rate, all plots in Figures (\ref{fig:performance_metrics_noise_10} and \ref{fig:fair_metrics_noise_10}) compare the behavior of GFLC (in red, dotted line) and Fair-OBNC (in blue, dashed line) as we vary the decision threshold from 0 to 0.6. Generally, these figures show results similar to those observed at a 5\% noise rate. In this context, we will skip detailing the results for the 10\% label noise rate and instead focus on the results for a 20\% label noise rate, as they exhibit more interesting behavior and highlight the advantages of our GFLC method for higher noise rates.

\subsection{Results at Noise Rate 20\%}

\begin{figure}[htbp]
\centering
\begin{subfigure}[b]{0.32\textwidth}
    \centering
    \includegraphics[width=0.95\textwidth]{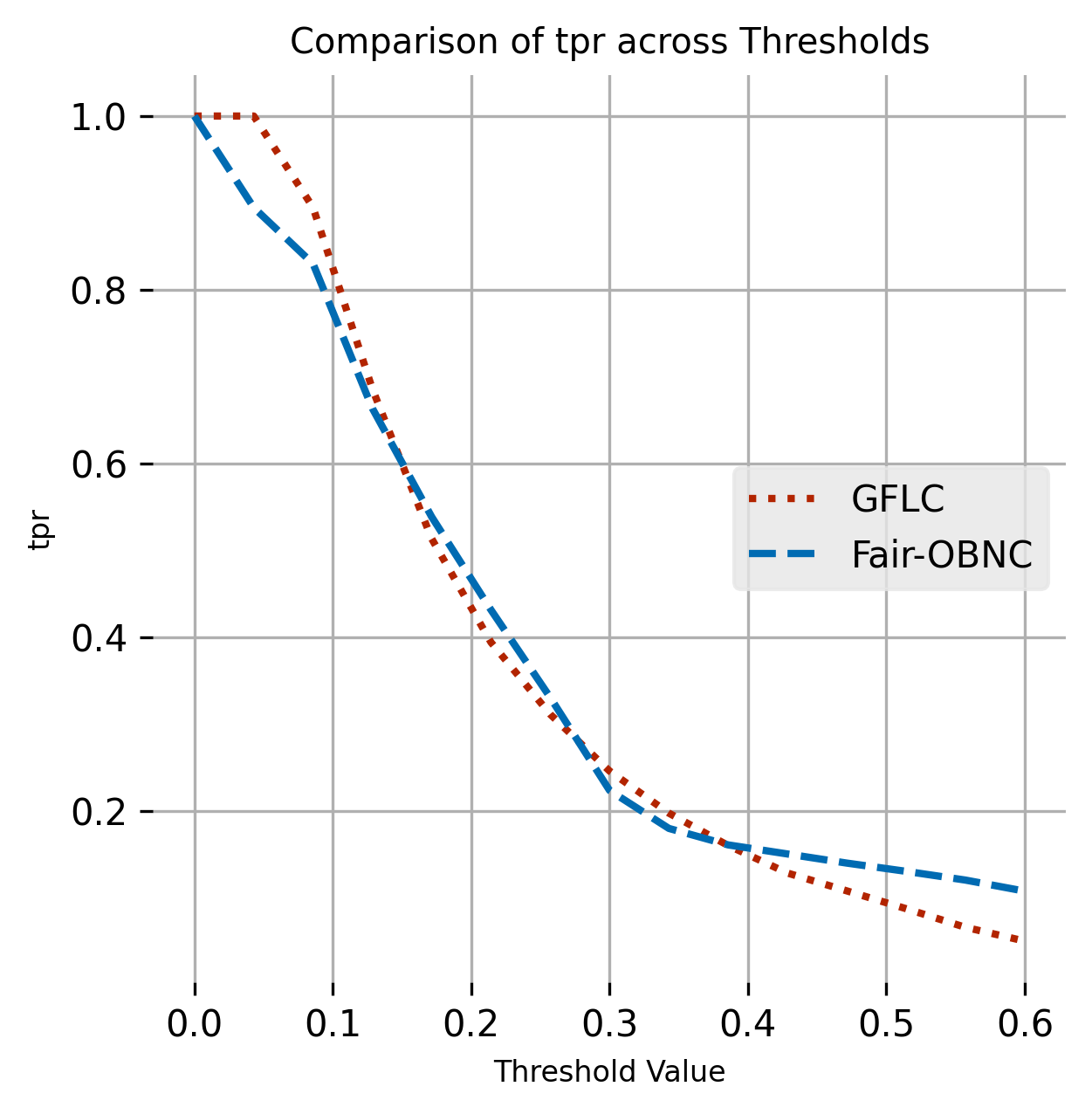}
    \caption{True Positive Rate (TPR)}
    \label{fig:tpr_noise_20}
\end{subfigure}
\hfill
\begin{subfigure}[b]{0.32\textwidth}
    \centering
    \includegraphics[width=0.95\textwidth]{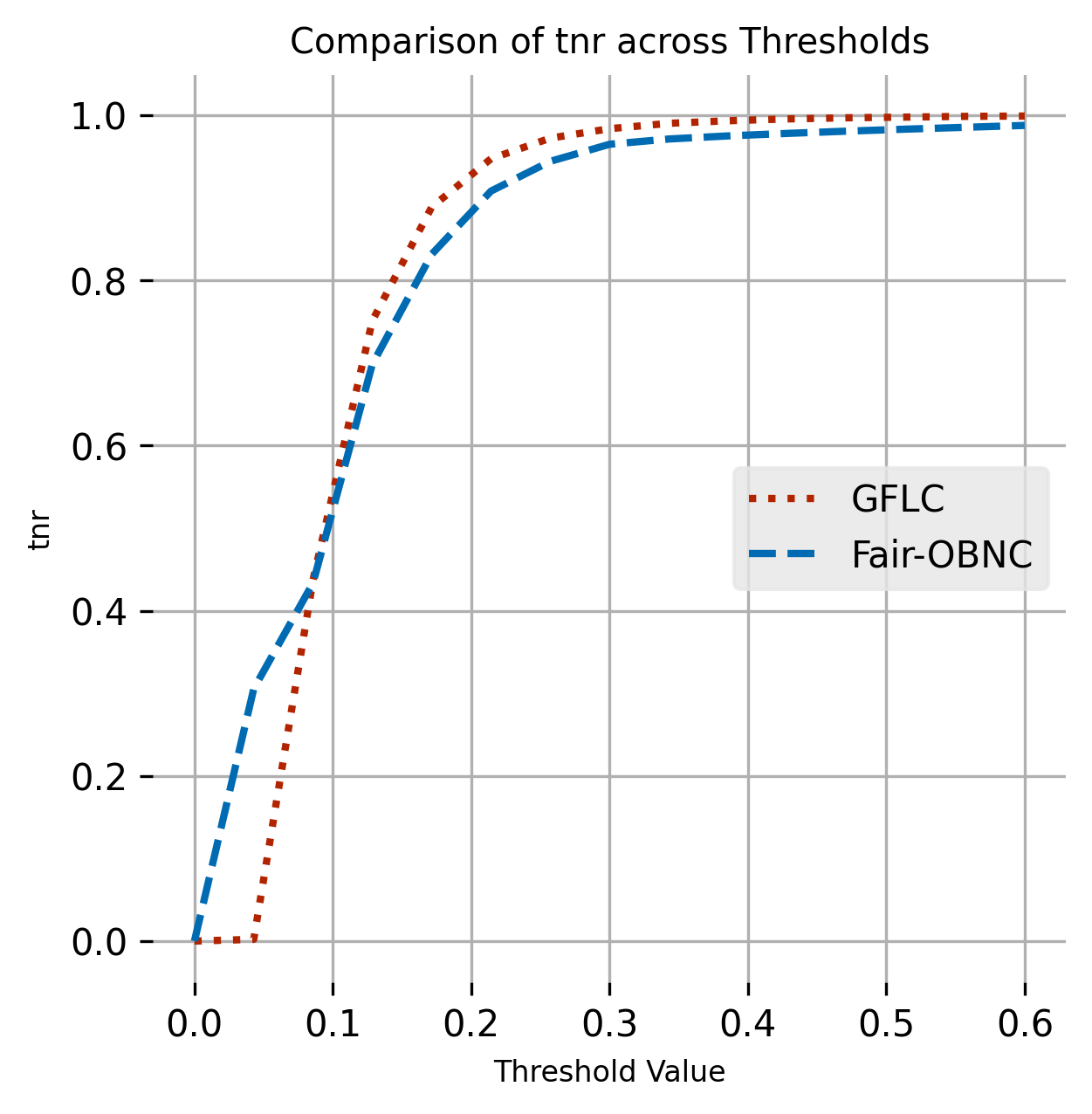}
    \caption{True Negative Rate (TNR)}
    \label{fig:tnr_noise_20}
\end{subfigure}
\hfill
\begin{subfigure}[b]{0.32\textwidth}
    \centering
    \includegraphics[width=0.95\textwidth]{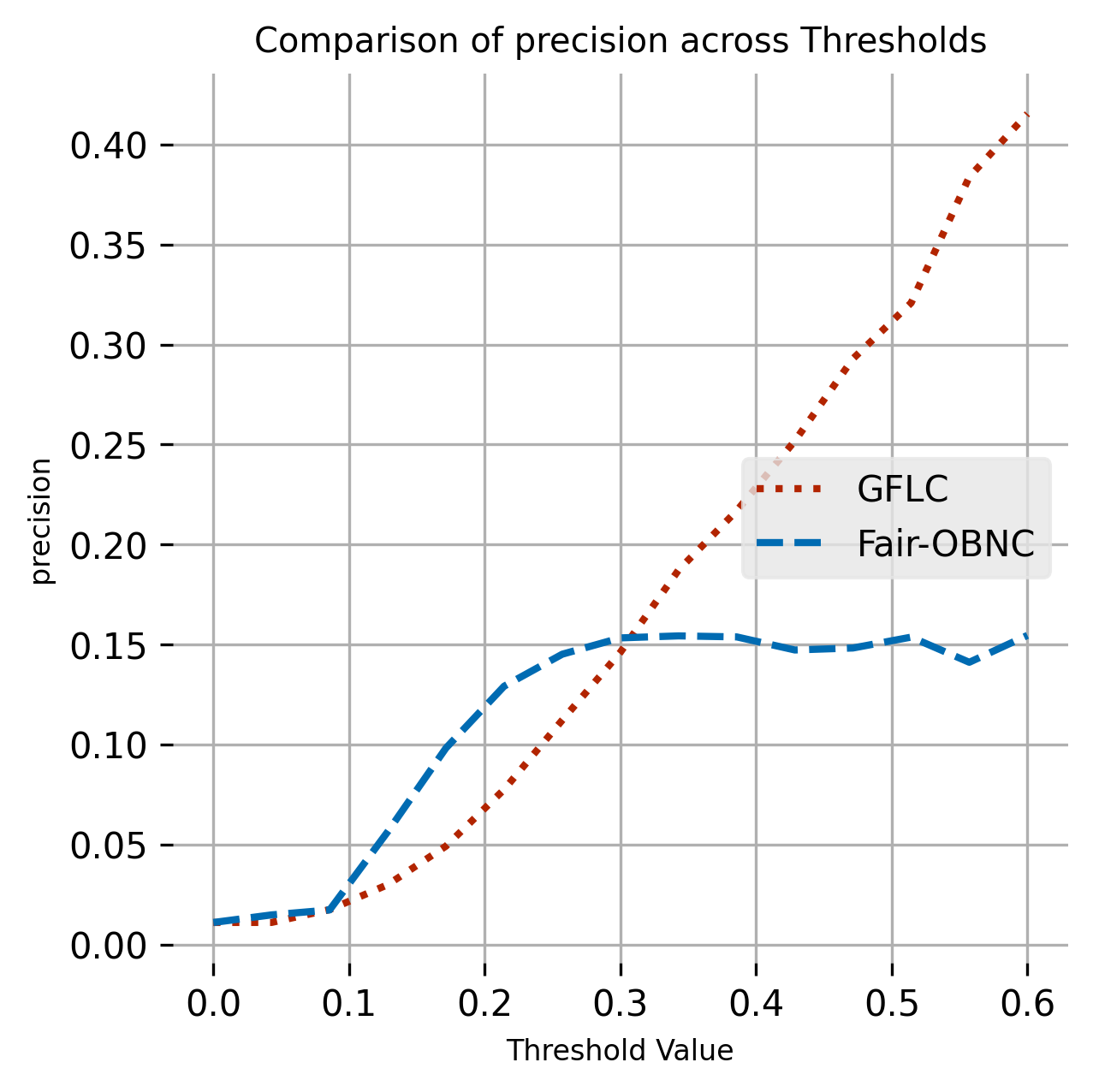}
    \caption{Precision}
    \label{fig:precision_noise_20}
\end{subfigure}

\vspace{0.2cm}

\begin{subfigure}[b]{0.32\textwidth}
    \centering
    \includegraphics[width=0.95\textwidth]{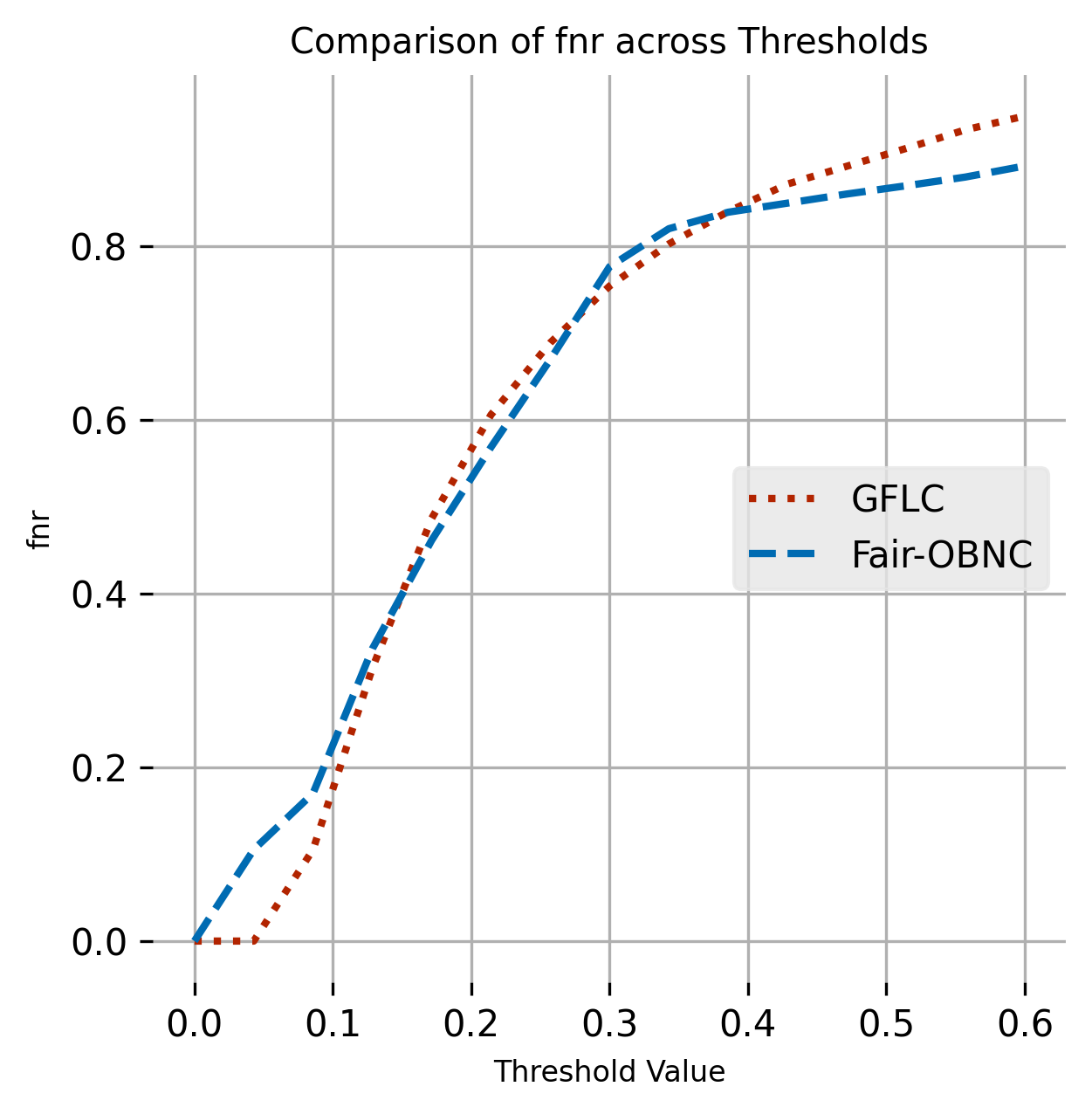}
    \caption{False Negative Rate (FNR)}
    \label{fig:fnr_noise_20}
\end{subfigure}
\hfill
\begin{subfigure}[b]{0.32\textwidth}
    \centering
    \includegraphics[width=0.95\textwidth]{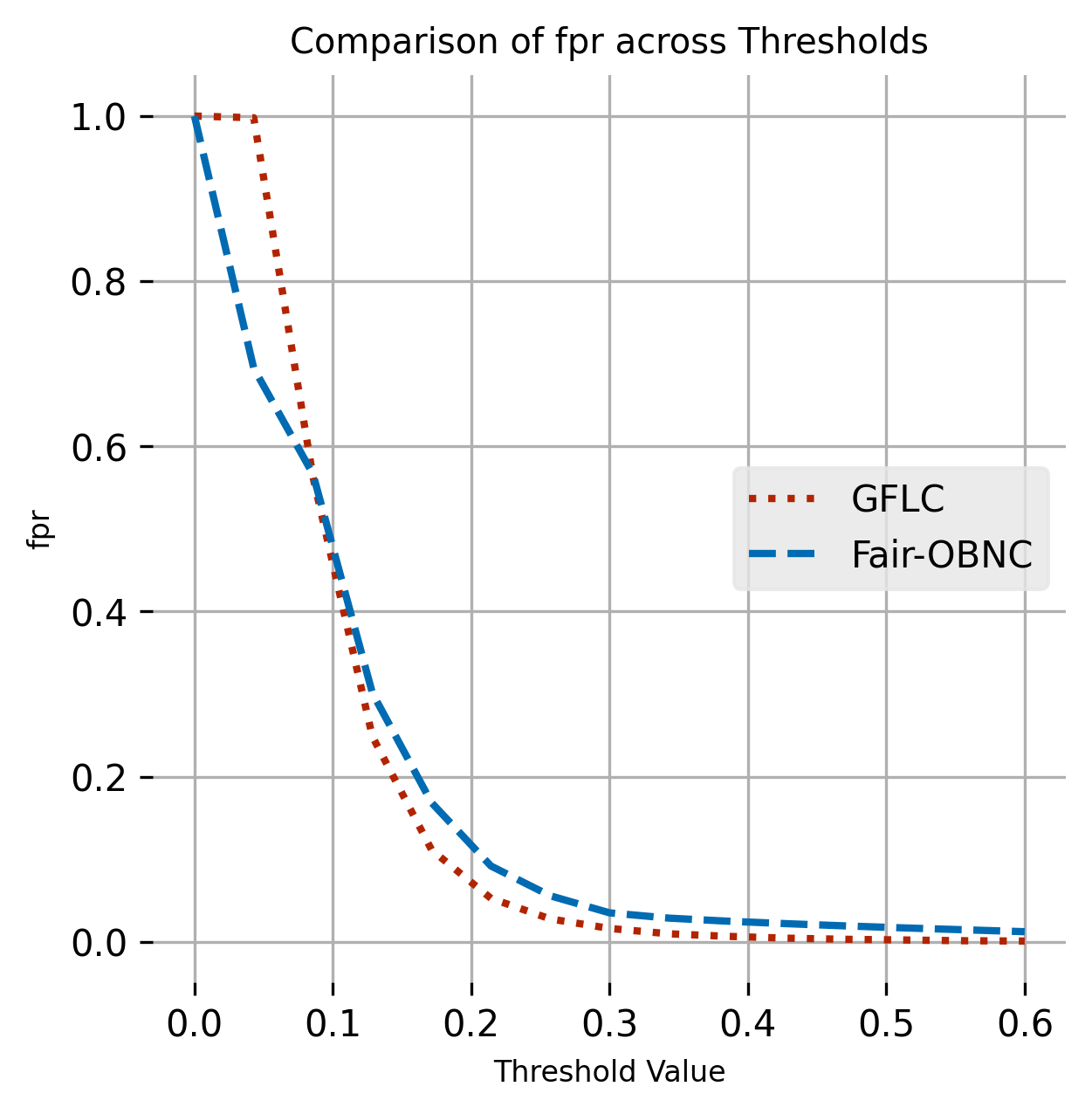}
    \caption{False Positive Rate (FPR)}
    \label{fig:fpr_noise_20}
\end{subfigure}

\caption{Performance Metrics Comparison Across Different Thresholds at 20\% Noise Rate}
\label{fig:performance_metrics_noise_20}
\end{figure}

\begin{figure}[htbp]
\centering
\begin{subfigure}[b]{0.32\textwidth}
    \centering
    \includegraphics[width=0.95\textwidth]{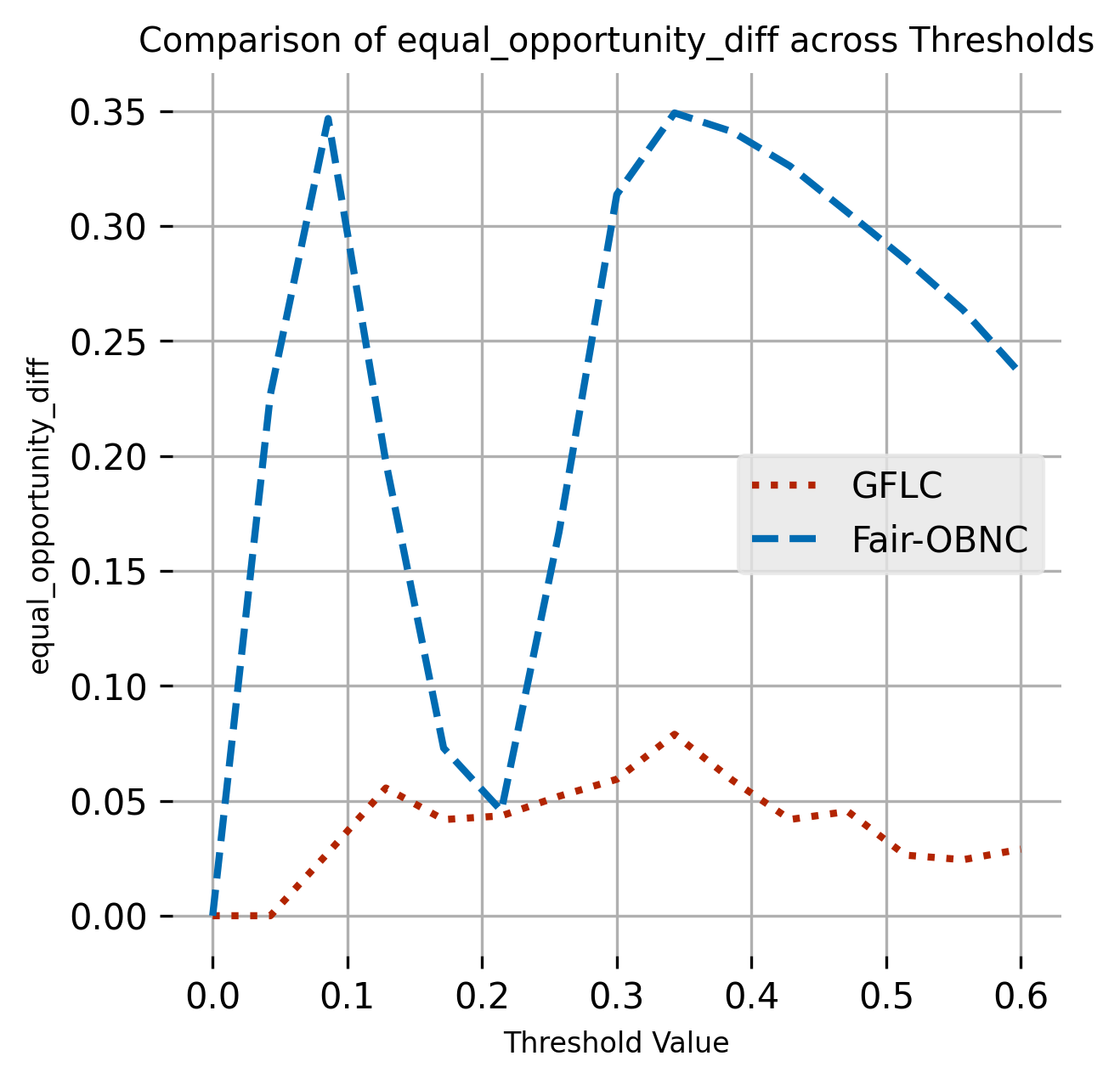}
    \caption{Equal Opportunity $\downarrow$}
    \label{fig:equal_opportunity_noise_20}
\end{subfigure}
\hfill
\begin{subfigure}[b]{0.32\textwidth}
    \centering
    \includegraphics[width=0.95\textwidth]{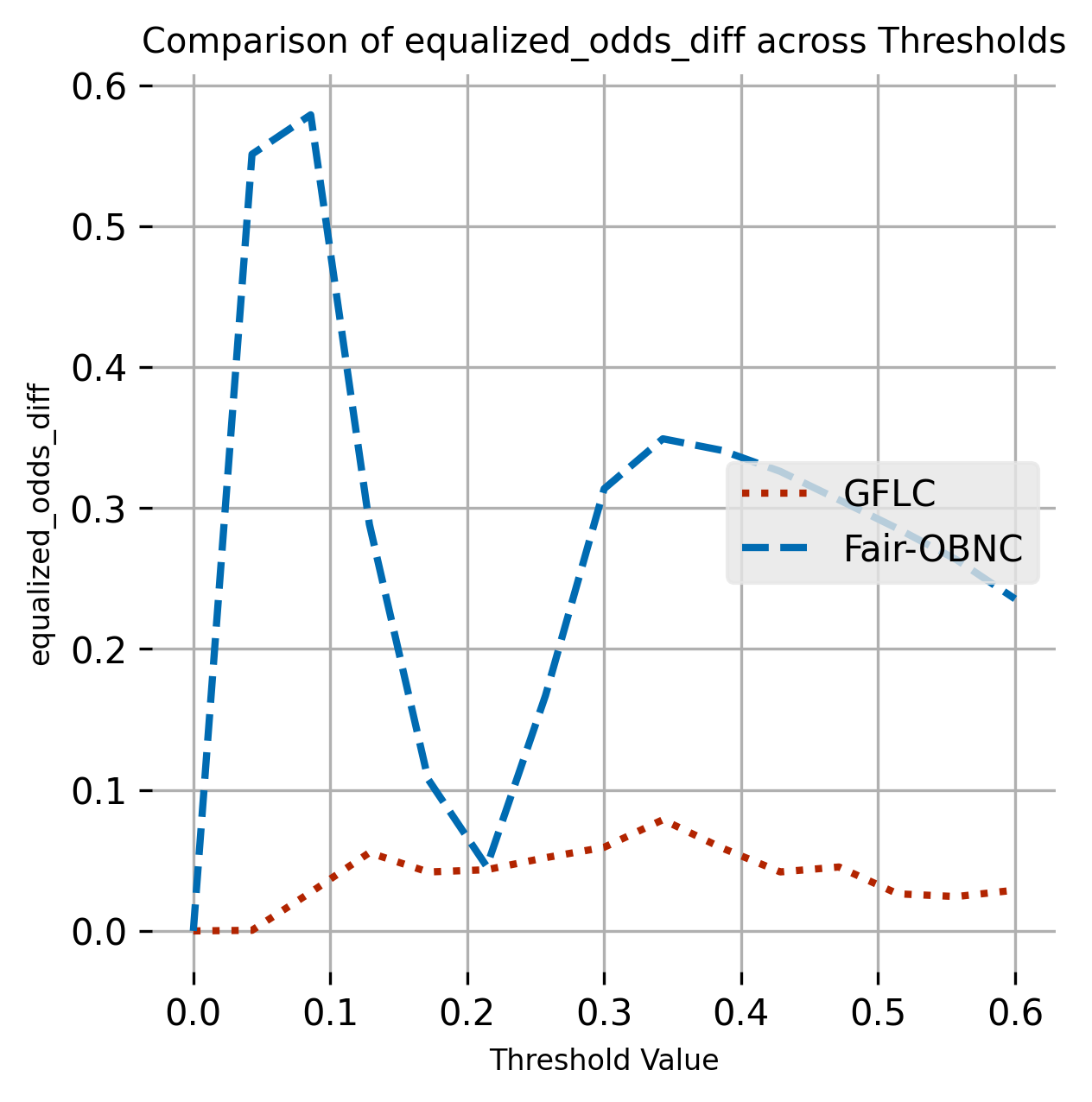}
    \caption{Equalized odds $\downarrow$}
    \label{fig:equalized_odds_noise_20}
\end{subfigure}
\hfill
\begin{subfigure}[b]{0.32\textwidth}
    \centering
    \includegraphics[width=0.95\textwidth]{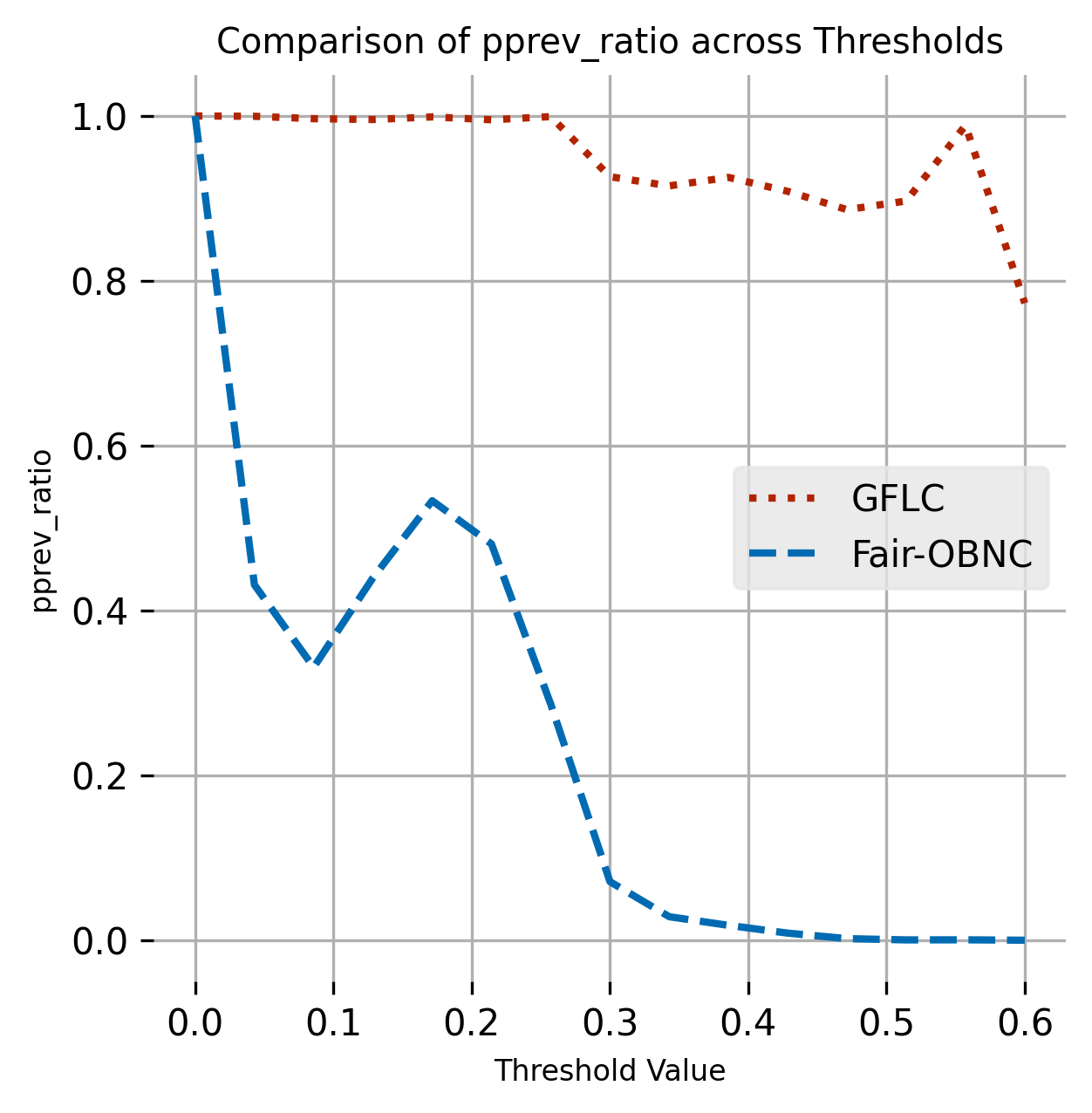}
    \caption{Demographic Parity $\uparrow$}
    \label{fig:pprev_ratio_noise_20}
\end{subfigure}

\caption{Fairness Metrics Comparison Across Different Thresholds at 20\% Noise Rate}
\label{fig:fair_metrics_noise_20}
\end{figure}

The plots in (Figure~\ref{fig:performance_metrics_noise_20} and Figure~\ref{fig:fair_metrics_noise_20}) compare how GFLC (red, dotted) and Fair‑OBNC (blue, dashed) behave as we change the decision threshold from 0 up to 0.6, at a 20\% label noise rate. The following analysis at a 20\% label noise rate, reveals the strength of our GFLC approach compared to the baseline method.

\paragraph{True‑Positive Rate (TPR) (Figure~\ref{fig:tpr_noise_20}):}

GFLC's TPR shows slightly higher values than Fair-OBNC's TPR as the threshold increases from 0 to 0.15. After the threshold of 0.15, Fair-OBNC's TPR remains slightly higher. This is opposite to the results observed at a 5\% label noise rate or a 10\% label noise rate, where GFLC's True Positive Rate (TPR) showed values very similar to Fair-OBNC's TPR as the threshold increased from 0 to 0.15. Furthermore, after the threshold exceeded 0.15, GFLC's TPR at the noise rate of 20\% exhibited behavior that was contrary to GFLC's TPR at the earlier noise rates of 5\% and 10\% and same as the Fair-OBNC.

\paragraph{False‑Positive Rate (FPR) (Figure~\ref{fig:fpr_noise_20}):}

As the threshold increases from 0 to 0.08, GFLC's FPR shows higher values than Fair‑OBNC. Between the thresholds of 0.08 and 0.11, both methods exhibit comparable behavior. However, after the threshold of 0.11, Fair‑OBNC's FPR shows higher values than GFLC.

\paragraph{Precision (Figure~\ref{fig:precision_noise_20}):} For low thresholds (less than 0.1) the Figure~\ref{fig:precision_noise_20} shows very low precision for both models again. In the mid-range (0.1–0.3), Fair-OBNC exhibits better precision values. However, beyond thresholds of 0.3, GFLC demonstrates superior precision compared to Fair-OBNC.

\paragraph{False Negative Rate (FNR):}

Figure~\ref{fig:fnr_noise_20} presents the false negative rate (FNR) across thresholds, where both methods show similar trends but with notable differences in magnitude. Both GFLC and Fair-OBNC demonstrate increasing FNR with higher thresholds. The convergence occurs around threshold 0.15, after which Fair-OBNC maintains higher FNR values.

\paragraph{True Negative Rate (TNR):}

Figure~\ref{fig:tnr_noise_20} demonstrates true negative rate (TNR) performance, where both methods achieve high performance levels above 0.95 for thresholds greater than 0.2. At lower thresholds ($ < 0.2$), GFLC shows slightly faster convergence to optimal TNR values.

\paragraph{Demographic Parity (Figure~\ref{fig:pprev_ratio_noise_20}):}

The demographic parity ratio shown in Figure~\ref{fig:pprev_ratio_noise_20} indicates that the GFLC method maintains exceptional demographic parity, with ratios consistently close to 1.0 across all threshold values. However, the Fair-OBNC method shows a significant degradation for the demographic parity metric as the thresholds increase, where it decreases from approximately 1.0 at a threshold of 0.0 to around 0 at a threshold of 0.6. This illustrates the advantages of the GFLC method, particularly at the high noise rates of 20 \% compared to Fair-OBNC.

\paragraph{Equal Opportunity and Equalized Odds:} Figures~\ref{fig:equal_opportunity_noise_20} and \ref{fig:equalized_odds_noise_20} show that our GFLC method outperforms the baseline method Fair-OBNC across all threshold values at this high noise rate of 20\% for equal opportunity and equalized odds. This highlights the advantages of our GFLC method for different fairness definitions in high-noise scenarios.

\section{Conclusion}\label{sec_Conclusion}

We introduced a novel approach for learning with instance-dependent noisy labels, called GFLC. Our method is developed to correct labels while ensuring fairness among different demographic groups. GFLC combines graph-based techniques, fairness-aware learning, and a prediction confidence measure. It effectively utilizes the structural insights provided by k-nearest neighbor (k-NN) graphs, discrete Forman–Ricci curvature, and discrete Ricci flow. Moreover, we demonstrated the effectiveness of our approach using a real dataset, showing that it improves the trade-off between discriminative performance and model fairness. This work builds on key aspects from the literature on instance-dependent noisy labels and highlights important issues related to fairness. Our work provides safer machine learning models that can be applied in significant societal contexts.








\bibliographystyle{dcu1}  
\bibliography{name}  

%

\appendix

\section{Derivation of Simplified Formula for Forman Curvature}\label{appendix_A}

Here, we present the derivation of the simplified Forman curvature formula described in Section \ref{Simplified_Forman}. This formula emphasizes edge-centric properties, indicating whether an edge serves as a bottleneck (negative curvature) or promotes expansion (positive curvature). We begin with the notion of discrete Ricci curvature for graphs or networks, namely, the Forman-Ricci curvature \citep{forman2003bochner, samal2018comparative}, and derive the simplified version through a series of justified steps. For an edge $e_{uv}$ connecting vertices $u$ and $v$, the original Forman curvature formula is as follows:

\begin{equation}
F(e_{uv}) = w_{uv}\left(\frac{w_u}{w_{uv}} + \frac{w_v}{w_{uv}} - \sum_{x \sim u} \frac{w_u}{\sqrt{w_{uv}w_{ux}}} - \sum_{x \sim v} \frac{w_v}{\sqrt{w_{uv}w_{vx}}}\right)
\end{equation}
where $w_{uv}$ is the edge weight between $u$ and $v$. Additionally, let $w_u$ and $w_v$ represent the weights, and let $x$ be the neighbors of vertex $u$. Furthermore, we assume unit vertex weights ($w_u = w_v = 1$), which is common in applications within network science. Thus, we obtain the following.

\begin{equation}
F(e_{uv}) = w_{uv}\left(\frac{2}{w_{uv}} - \sum_{x \sim u} \frac{1}{\sqrt{w_{uv}w_{ux}}} - \sum_{x \sim v} \frac{1}{\sqrt{w_{uv}w_{vx}}}\right)
\end{equation}

\begin{align}
= w_{uv}\left(\frac{2}{w_{uv}} - \frac{1}{w_{uv}^{1/2}} \left[\sum_{x \sim u} \frac{1}{\sqrt{w_{ux}}} + \sum_{x \sim v} \frac{1}{\sqrt{w_{vx}}}\right] \right)
\end{align}

Next, we will divide the entire equation by 2.

\begin{align}
\frac{F(e_{uv})}{2} &= \frac{w_{uv}}{2}\left(\frac{2}{w_{uv}} - \frac{1}{w_{uv}^{1/2}} \left[\sum_{x \sim u} \frac{1}{\sqrt{w_{ux}}} + \sum_{x \sim v} \frac{1}{\sqrt{w_{vx}}}\right]   \right) \\
&= 1 - \frac{w_{uv}^{1/2}}{2} \left[\sum_{x \sim u} \frac{1}{\sqrt{w_{ux}}} + \sum_{x \sim v} \frac{1}{\sqrt{w_{vx}}}\right]
\end{align}

After that, we define the normalized curvature \( F_{\text{norm}}(e_{uv}) \).

    \begin{equation}
    F_{\text{norm}}(e_{uv}) := \frac{F(e_{uv})}{2} = 1 - \frac{1}{2}\sqrt{w_{uv}} \left[\sum_{x \sim u} \frac{1}{\sqrt{w_{ux}}} + \sum_{x \sim v} \frac{1}{\sqrt{w_{vx}}}\right]
    \end{equation}

\textit{Final Adjustment for GFLC}: We apply the \( w_{uv} \) multiplication to the normalized curvature \( F_{\text{norm}}(e_{uv}) \) to ensure that strong edges dominate curvature calculations while weak edges in sparse areas do not appear artificially significant.

\begin{equation}
F_{\text{final}}(e_{uv}) = w_{uv} \cdot F_{\text{norm}}(e_{uv}) 
\end{equation}

\begin{align}
F_{\text{final}}(e_{uv}) &= w_{uv} \cdot F_{\text{norm}}(e_{uv}) \\
&= w_{uv}\left(1 - \frac{1}{2}\sqrt{w_{uv}}   \left[\sum_{x \sim u} \frac{1}{\sqrt{w_{ux}}} + \sum_{x \sim v} \frac{1}{\sqrt{w_{vx}}}\right]  \right)
\end{align}

\end{document}